\documentclass[11pt]{article}

\usepackage[preprint]{acl}

\usepackage{times}
\usepackage{latexsym}

\usepackage[T1]{fontenc}

\usepackage[utf8]{inputenc}

\usepackage{microtype}

\usepackage{inconsolata}

\usepackage{graphicx}
\usepackage{listings}
\usepackage{multirow}
\usepackage{tcolorbox}
\tcbuselibrary{breakable, listings}

\usepackage{listings}
  \lstdefinelanguage{plain}{}  
  \lstdefinestyle{grayoutput}{
    language=plain,
    basicstyle=\scriptsize\ttfamily\color{gray},
    identifierstyle=\color{gray},
    breaklines=true,
    columns=fullflexible,
    breakautoindent=false,
    frame=none,
    xleftmargin=0.5em,
    aboveskip=0.5ex,
    belowskip=0.5ex
  }
\lstdefinelanguage{plain}{}
  \lstdefinestyle{grayoutput}{
    language=plain,
    basicstyle=\scriptsize\ttfamily\color{gray},
    identifierstyle=\color{gray},
    breaklines=true,
    columns=fullflexible,
    breakautoindent=false,
    frame=none,
    xleftmargin=0.5em,
    aboveskip=0.5ex,
    belowskip=0.5ex,
    literate=
      {\#}{{\textcolor{gray}{\#}}}1
      {[}{{\textcolor{gray}{[}}}1
      {]}{{\textcolor{gray}{]}}}1
      {-}{{\textcolor{gray}{-}}}1
      {:}{{\textcolor{gray}{:}}}1
      {(}{{\textcolor{gray}{(}}}1
      {)}{{\textcolor{gray}{)}}}1
      {/}{{\textcolor{gray}{/}}}1
      {.}{{\textcolor{gray}{.}}}1
      {,}{{\textcolor{gray}{,}}}1
      {;}{{\textcolor{gray}{;}}}1
      {\_}{{\textcolor{gray}{\_}}}1
      {"}{{\textcolor{gray}{"}}}1
      {0}{{\textcolor{gray}{0}}}1
      {1}{{\textcolor{gray}{1}}}1
      {2}{{\textcolor{gray}{2}}}1
      {3}{{\textcolor{gray}{3}}}1
      {4}{{\textcolor{gray}{4}}}1
      {5}{{\textcolor{gray}{5}}}1
      {6}{{\textcolor{gray}{6}}}1
      {7}{{\textcolor{gray}{7}}}1
      {8}{{\textcolor{gray}{8}}}1
      {9}{{\textcolor{gray}{9}}}1
  }
\newtcbox{\code}{on line, boxrule=0pt, boxsep=0pt,colback=gray!15, top=2pt, bottom=2pt, left=3pt, right=3pt,arc=2pt,fontupper=\ttfamily}
\usepackage{mdframed}
\usepackage{xcolor}
\usepackage{enumitem}
\setlist{itemsep=0pt, parsep=0pt, topsep=2pt}
\usepackage{booktabs}
\usepackage{comment}
\usepackage{amssymb}
\usepackage{tabularx}
\usepackage{floatrow}
\usepackage[table]{xcolor}
\newcommand{\sysname}{MemCog}
\newcommand{\memstruc}{Navigable Memory Store}
\newcommand{\navimemstruc}{navigable memory structure}
\title{\sysname{}: From Memory-as-Tool to Memory-as-Cognition in Conversational Agents}

\author{
  \textbf{Zihan Li\textsuperscript{1}},
  \textbf{Xingyu Fan\textsuperscript{1}},
  \textbf{Feifei Li\textsuperscript{1}},
  \textbf{Wenhui Que\textsuperscript{1}}
\\
  \textsuperscript{1}WeChat, Tencent Inc., Beijing, China
\\
  \texttt{\{muselli, fanxfan, niyali, victorque\}@tencent.com}
}

\begin{document}
\maketitle
\begin{abstract}
Existing agent memory systems universally follow what we term a Memory-as-Tool paradigm where a single query triggers one-shot retrieval of flat passage lists, suffering from passive invocation, reasoning-retrieval decoupling, and structural mismatch between retrieved fragments and the agent's navigational needs. We propose \textbf{\sysname{}}, a \textbf{Memory-as-Cognition} system that makes memory access an integral part of the reasoning process. \sysname{} organizes user knowledge as \memstruc{} with associative link graphs, exposes Cross-Dimensional Navigation Interface for multi-step reasoning-driven traversal, and employs Proactive Reasoning Protocol that drives agents to spontaneously initiate memory exploration from conversational context. We additionally construct ProactiveMemBench\footnote{Available at \url{https://anonymous.4open.science/r/ProactiveMemBench-76E1/}}, the first benchmark for evaluating proactive memory triggering. Experiments show that \sysname{} achieves state-of-the-art on passive QA benchmarks (92.98 on LoCoMo, 95.8 on LongMemEval) while substantially outperforming baselines on ProactiveMemBench, demonstrating the advantage of Memory-as-Cognition.

\end{abstract}

\section{Introduction}
\label{sec:intro}

Long-term memory \cite{memgpt,mem0,zep,amem,hypermem} is fundamental to maintaining coherent, personalized interactions across extended dialogues, enabling agents to capture evolving user knowledge from fragmented conversations and generate historically informed responses beyond fixed context windows. From MemGPT\cite{memgpt} to HyperMem\cite{hypermem}, agent memory systems have grown increasingly sophisticated in \textit{what} they retrieve. Yet they share a common usage pattern: a single query produces a flat list of passages, and the agent reasons only after retrieval concludes. We argue that this pattern, which we term \textbf{Memory-as-Tool}, imposes fundamental limitations on how agents can utilize long-term knowledge.

\begin{figure}[t]
    \centering
    \includegraphics[width=\columnwidth]{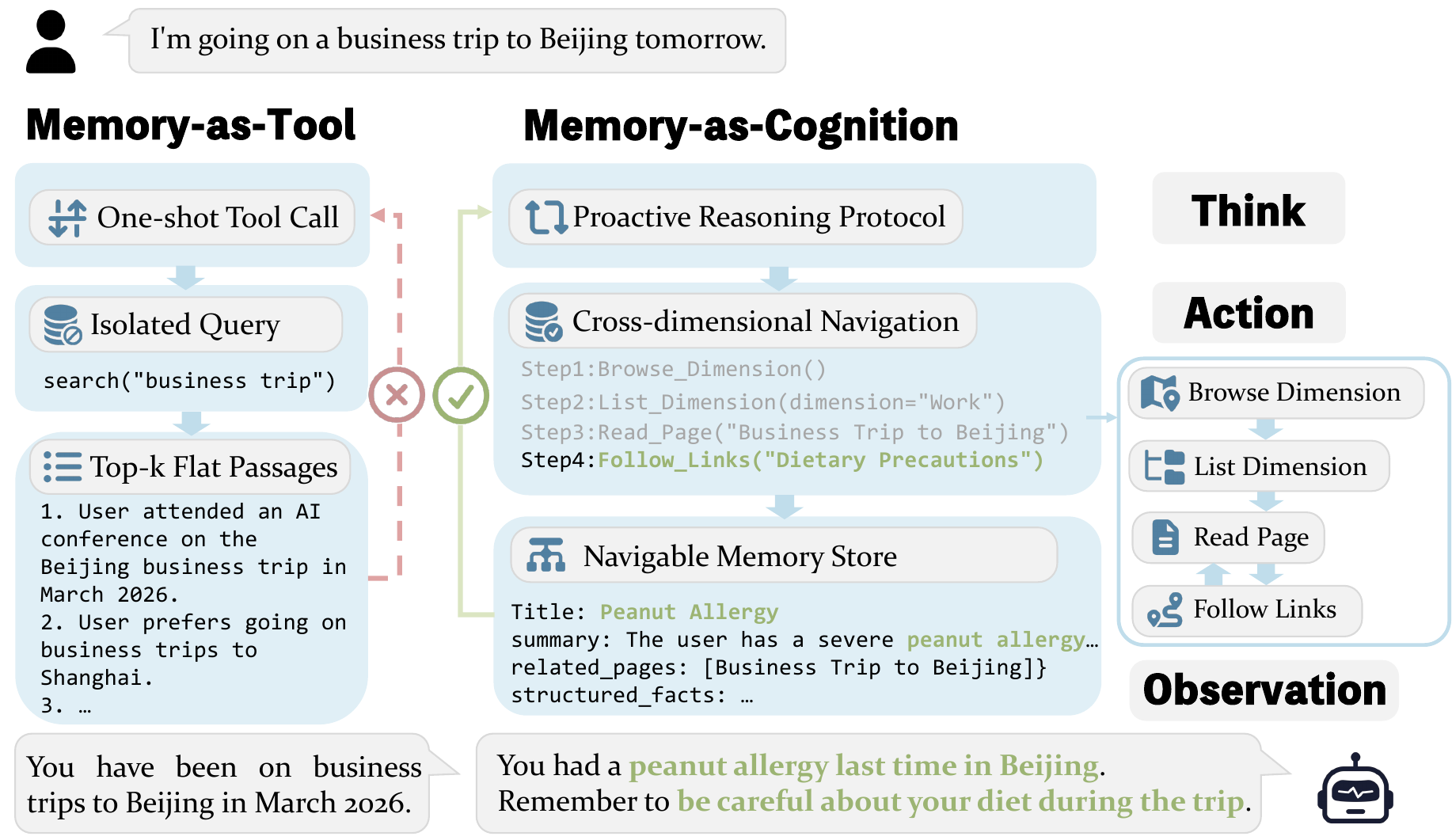}
    \caption{Comparison of Memory-as-Tool vs Memory-as-Cognition patterns.}
    \label{fig:memory-comparison}
\end{figure}

This Memory-as-Tool paradigm imposes three fundamental limitations. First, \textbf{Invocation Bottleneck}: the memory system is activated only when the agent explicitly issues a tool call to retrieve user memories relevant to a question, leaving memory as a passive responder to query demands. The memory system lacks the ability to proactively trigger retrieval. 
Existing benchmarks evaluate exclusively around factual questions, diverging from daily conversational scenarios and the proactive triggering patterns of human memory.

Second, \textbf{Reasoning-Retrieval Decoupling}: 
the Memory-as-Tool system retrieval remains a one-shot call, preventing the agent from dynamically planning further exploration when initial results are insufficient. The agent's reasoning capabilities cannot intervene in the memory system's isolated retrieval pipeline, which constrains response quality and forces the two modules to optimize independently without co-evolution.

Third, \textbf{Structural Mismatch}: 
Even when memory is organized hierarchically\cite{amem,zep,hypermem}, the top-$k$ fragments returned by retrieval lack structural navigational information, preventing agents from inferring optimal next-step retrieval paths from current results. Existing methods cannot extend their structural capabilities to guide the agent's directed iterative retrieval within the memory store.

To address the above three limitations, we propose \sysname{}, the first memory system that transitions from Memory-as-Tool to \textbf{Memory-as-Cognition}. First, we design a \textbf{Proactive Reasoning Protocol} that drives agents from passively waiting for queries to actively reasoning within memory, addressing the invocation bottleneck. 
We advance \textbf{toward} a Memory-as-Cognition paradigm in which retrieval and reasoning become inseparable within a ReAct loop augmented by proactive memory triggering. This design serves as a principled engineering stepping stone, anticipating the eventual maturation of fully architectural solutions for deployment in personalized agent settings.
Second, we instantiate a \textbf{Cross-Dimensional Navigation Interface} that replaces the single retrieval tool call with multi-granularity navigation actions, where the agent autonomously determines the next action at each step based on current observations, achieving tight coupling between retrieval and reasoning. Third, we propose \textbf{\memstruc{}}, which provides agents with structured navigational paths from hierarchical dimensions to specific memory sections, enabling agents to plan subsequent navigation based on the structural context returned at each step.
Additionally, we construct \textbf{ProactiveMemBench}, the first benchmark for evaluating proactive memory triggering capabilities, filling the gap left by existing factual QA benchmarks limited to passive retrieval scenarios.


To the best of our knowledge, \sysname{} is the first system that unifies proactive reasoning with structured navigational memory, transforming memory access from a passive preprocessing step into an integral part of the agent's reasoning process. Our main contributions are summarized as follows:
\begin{itemize}
    \item We propose \textbf{Memory-as-Cognition}, a paradigm that transforms passive memory access into an integral part of the agent's reasoning process
    , thereby eliminating the invocation bottleneck inherent in Memory-as-Tool systems.

    \item We construct \textbf{ProactiveMemBench}, the first benchmark for evaluating proactive memory triggering, built through a six-step LLM-driven pipeline spanning 500 instances across five topic domains with 98.4\% human validation acceptance.

    \item Experiments on three benchmarks validate the effectiveness of \sysname{}, achieving state-of-the-art on passive QA tasks and substantially outperforming baselines on proactive memory triggering. Ablation studies further confirm the protocol–structure synergy in \sysname{}, with all components proving indispensable and mutually reinforcing.
\end{itemize}

\section{Related Work}

\subsection{Single-Shot Memory Retrieval}

The most prevalent paradigm treats retrieval as a single-round tool call. Mem0 \cite{mem0} retrieves atomic facts via semantic similarity; MemGPT \cite{memgpt} swaps fixed-size pages through function calls inspired by OS virtual memory; LangMem \cite{langmem} provides an SDK-level search interface. Despite different storage backends, these systems share a common access pattern: one query in, one flat list out, with no opportunity for iterative refinement.

\subsection{Structured Memory Organization}

Recent work enriches memory with structural priors while retaining the single-shot interface. HyperMem \cite{hypermem} employs a hypergraph-based three-level funnel capturing high-order associations, yet executes as a single pass. A-Mem \cite{amem} adopts Zettelkasten-inspired notes with activation-based triggering but returns results in one batch. HippoRAG \cite{hipporag} builds a knowledge graph with Personalized PageRank retrieval in a single step. These innovations improve \textit{what} is retrieved but leave the query-in, passages-out access pattern unchanged.

\subsection{Multi-Step Reasoning over Knowledge}
\begin{figure*}[t]
    \centering
  \includegraphics[width=\textwidth, trim=0cm 0cm 6.7cm 0cm, clip]{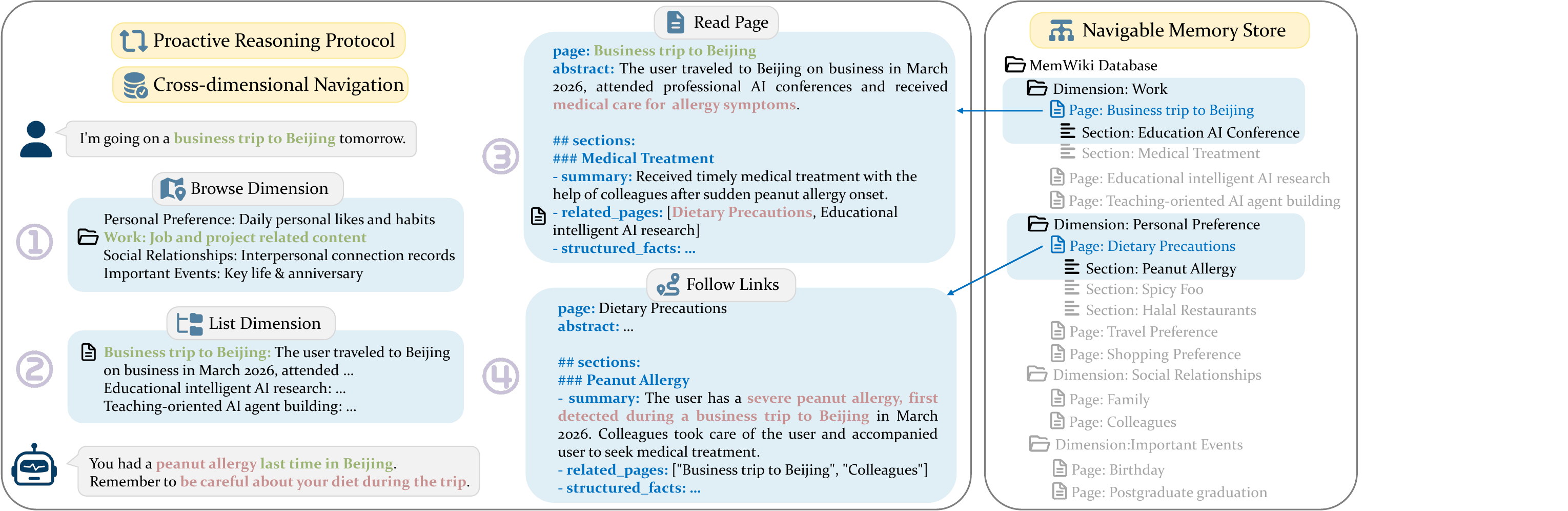}
    \caption{Example of \sysname{}: Proactive Reasoning Protocol: \memstruc{} and Cross-Dimensional Navigation path examples.}
    \label{fig:\sysname{}-protocol}
\end{figure*}
A parallel line explores multi-step retrieval coupled with reasoning. MemWalker \cite{memwalker} navigates summary trees via zero-shot prompting but targets single-document comprehension; RAPTOR \cite{raptor} builds static hierarchical trees offline without interactive navigation; IRCoT \cite{ircot} interleaves retrieval with chain-of-thought but lacks persistent user memory; Openhuman \cite{openhuman} applies wiki-style organization to document corpora. \sysname{} instead targets conversational memory with proactive triggering as the core innovation.


\subsection{Proactive Systems}
Existing proactive systems operate at dialogue strategy level rather than memory triggering: proactive dialogue systems \cite{proactivedialogue} steers conversations toward predefined goals; Generative Agents \cite{generativeagents} synthesize memories via scheduled reflection; 
Agentic Retrieval-Augmented Generation \cite{adaptiverag,crag,speculativerag} improve retrieval strategies but involve no agent-driven navigation. \sysname{}'s proactivity instead operates at the memory level: agents spontaneously initiate multi-step navigation when contextual cues suggest relevant stored associations.


\section{Method}
\subsection{Overview}

As illustrated in Figure~\ref{fig:\sysname{}-protocol}, \sysname{} comprises three tightly integrated components: (1) \textbf{\memstruc{}} (\S\ref{sec:navi-structure}) that organizes user knowledge into a hierarchical, link-rich database; (2) \textbf{Cross-Dimensional Navigation Interface} (\S\ref{sec:navigation}) that replaces single-shot retrieval with multi-step, reasoning-driven memory traversal; (3) \textbf{Proactive Reasoning Protocol} (\S\ref{sec:protocol}) that drives agents to spontaneously initiate memory exploration during real-time conversation. Together, these components transform memory access from the conventional Memory-as-Tool pattern (one query in, flat passages out) into a \textbf{Memory-as-Cognition} process where retrieval and reasoning are interleaved across multiple steps, with each navigation action informed by prior observations.

\subsection{\memstruc{}}
\label{sec:navi-structure}

\subsubsection{Hierarchical Organization}

The core design principle of \memstruc{} is to support agent multi-step navigation rather than optimizing for single-shot retrieval. We organize user knowledge as a hierarchical memory store with three levels of granularity:

\begin{itemize}
    \item \textbf{Dimension}: top-level categories that partition the user's knowledge space (e.g., personal preferences, work, social relationships, important events).
    \item \textbf{Page}: individual knowledge units within each dimension, each covering a coherent topic.
    \item \textbf{Section}: fine-grained information blocks within a page, containing title, summary, structured facts, related pages, temporal context, and confidence scores.
\end{itemize}

Dimensions provide entry points, Pages provide depth, and cross-dimensional associative links within Sections provide lateral connectivity. This design uniquely combines hierarchical browsability with cross-branch linkage, ensuring that every step returned during navigation carries sufficient structural context for the agent to reason about what to explore next.

\subsubsection{Cross-dimensional associative Link}

Cross-dimensional associative links establish an \textbf{overlay graph} that facilitates seamless traversal across dimensional boundaries. We define four link types to support different reasoning patterns:

\begin{itemize}
    \item \textbf{related\_to}: semantic associations across dimensions (e.g., ``Dietary Precautions'' $\leftrightarrow$ ``Business Trip to Beijing''), supporting cross-dimensional association.
    \item \textbf{temporal\_next}: directed temporal succession (e.g., ``project v1'' $\rightarrow$ ``project v2''), supporting timeline tracing.
    \item \textbf{caused\_by}: directed causal relations (e.g., ``peanut allergy'' $\leftarrow$ ``restaurant incident''), supporting causal reasoning.
    \item \textbf{contrasts\_with}: contrastive relations (e.g., ``like spicy'' $\leftrightarrow$ ``dislike sweet''), supporting preference comparison.
\end{itemize}

These typed links serve as explicit navigational affordances: when an agent reads a page, the visible links suggest further exploration directions, naturally inducing multi-step traversal without requiring additional prompting.

\subsubsection{Construction and Evolution}

\paragraph{Initial Construction.} Given a user's conversation history, Page construction proceeds in three stages: (1) an LLM extracts user-relevant information from conversations as structured facts; (2) facts are clustered into dimensions and pages via semantic grouping; (3) pages are organized into sections, which are interconnected via cross-dimensional associative links established based on entity co-occurrence, temporal proximity, and causal indicators.

\paragraph{Incremental Update.} After each conversation turn, Page evolves through: (1) extracting new information from the latest dialogue; (2) matching extracted facts to existing pages for section updates, or creating new pages when no match is found; (3) establishing cross-links for newly created or substantially updated pages; (4) performing contradiction detection and confidence score updates when new information conflicts with existing entries.

\subsection{Cross-Dimensional Navigation}
\label{sec:navigation}


Rather than exposing a single search tool that returns top-$k$ fragments for a given query, \sysname{} provides agents with a set of navigation actions at multiple granularities:

\begin{itemize}
    \item \textbf{list\_dimensions()}: returns all top-level dimensions with brief descriptions, providing a global overview of the memory store.
    \item \textbf{browse\_dimension(dim)}: returns page titles and summaries within a specified dimension, enabling targeted exploration.
    \item \textbf{read\_page(page\_id)}: returns the full content of a specific page including all sections and outgoing links.
    \item \textbf{follow\_link(link)}: navigates an inter-page hyperlink and returns the target page content, enabling cross-dimensional traversal without requiring the agent to know the top-level structure of other dimensions.
\end{itemize}

This design offers agents multiple granularities of access. The agent autonomously decides which action to invoke at each step based on its current reasoning state. Critically, each action returns not only the requested content but also structural context (available links, sibling pages, dimensional position) that informs subsequent navigation decisions.

\subsection{Proactive Reasoning Protocol}
\label{sec:protocol}

\paragraph{Protocol Design.}
The Proactive Reasoning Protocol is implemented as a structured system prompt defining the agent's memory-usage behavioral norms (Appendix~\ref{sec:appendix}). It specifies two modes of memory engagement: \textbf{(1) Proactive association}: 
the agent spontaneously initiates memory navigation when conversational context suggests potential connections to stored knowledge, 
specifically when: (i) mentions of people, places, or topics link to past memories; (ii) the conversation would benefit from historical context; (iii) user statements contradict stored information.
\textbf{(2) Reactive response}: the agent navigates memory in response to explicit user questions about past information. 

The protocol further specifies cross-dimensional navigation rather than stopping at first results, following links to build complete pictures, integrating information naturally, adopting autonomous routing strategies, and reasoning about which dimensions are most relevant before acting.

\paragraph{Selective Surfacing.}
Proactive retrieval does not imply indiscriminate information sharing. The agent evaluates recalled information along four dimensions before surfacing: relevance to current context, temporal validity, sensitivity, and redundancy. These judgments are performed by the LLM's own reasoning during the Think phase, requiring no additional gating module.
Detailed strategies and illustrative examples are provided in Appendix~\ref{appendix_navigation} and Table~\ref{tab:navigation_strategies}.






\subsection{ProactiveMemBench}
\begin{table*}[t]
\centering
\caption{Comparison of \sysname{} with Memory-as-Tool methods on the LoCoMo benchmark. The QA and LLM-as-Judge configurations follow \cite{mem0}. $^\dagger$ indicates results sourced from \cite{memr3} and \cite{hypermem}; all other baselines are reproduced using their official implementations. Best results in \textbf{bold}, second best \underline{underlined}. \colorbox{gray!12}{Gray} rows denote our full system. The same convention applies to all subsequent tables.}
\label{tab:locomo}
\resizebox{0.75\textwidth}{!}{
\begin{tabular}{@{}c||l|ccccc@{}}
\toprule
\textbf{Model} & \textbf{Method} & \textbf{Single-hop} & \textbf{Temporal} & \textbf{Multi-hop} & \textbf{Open-ended} & \textbf{Overall} \\
\midrule
\multirow{8}{*}{\rotatebox{90}{GLM-5.1}} & Full Context & 69.50 & 38.75 & 51.04 & 89.66 & 72.97 \\
& mem0\cite{mem0} & 46.75 & 76.56 & 57.29 & 80.74 & 56.16 \\
& A-Mem\cite{amem} & 48.94 & 27.81 & 36.46 & 72.06 & 56.40 \\
& ZEP\cite{zep} & 42.55 & 31.56 & 41.67 & 46.25 & 33.25 \\
& HyperMem\cite{hypermem} & \textbf{85.99} & 88.75 & 67.71 & \textbf{93.94} & \underline{89.77} \\
\cmidrule{2-7}
& \sysname{} w/o Proactive & 84.52 & \textbf{92.23} & \underline{73.03} & 91.91 & 89.48 \\
& \sysname{} w/o Graph Overlay & 78.37 & 88.13 & 70.83 & 89.30 & 85.90 \\
& \cellcolor{gray!12}\textbf{\sysname{} (full)} & \cellcolor{gray!12}\underline{85.28} & \cellcolor{gray!12}\underline{91.56} & \cellcolor{gray!12}\textbf{76.04} & \cellcolor{gray!12}\underline{92.27} & \cellcolor{gray!12}\textbf{89.83} \\
\midrule
\multirow{8}{*}{\rotatebox{90}{GPT-4o-mini}} & Full Context$^\dagger$ & 67.38 & 73.83 & 63.54 & 78.67 & 74.62 \\
& mem0$^\dagger$\cite{mem0} & 67.13 & 55.51 & 51.15 & 72.93 & 66.88 \\
& A-Mem$^\dagger$\cite{amem} & 61.70 & 64.49 & 40.62 & 76.63 & 69.06 \\
& ZEP$^\dagger$\cite{zep} & 67.38 & 73.83 & 63.54 & 78.67 & 74.62 \\
& HyperMem\cite{hypermem} & 71.28 & 85.00 & 55.21 & 90.61 & 83.69 \\
\cmidrule{2-7}
& \sysname{} w/o Proactive & \textbf{90.44} & \underline{92.18} & \textbf{81.71} & \underline{93.33} & \underline{91.84} \\
& \sysname{} w/o Graph Overlay & 86.52 & 90.62 & 77.08 & 92.98 & 90.32 \\
& \cellcolor{gray!12}\textbf{\sysname{} (full)} & \cellcolor{gray!12}\underline{90.43} & \cellcolor{gray!12}\textbf{93.13} & \cellcolor{gray!12}\underline{80.21} & \cellcolor{gray!12}\textbf{94.65} & \cellcolor{gray!12}\textbf{92.66} \\
\midrule
\multirow{8}{*}{\rotatebox{90}{GPT-4.1-mini}} & Full Context$^\dagger$ & 86.43 & 86.82 & 71.88 & 93.73 & 89.00 \\
& mem0$^\dagger$\cite{mem0} & 62.41 & 57.32 & 44.79 & 66.47 & 62.47 \\
& A-Mem$^\dagger$\cite{amem} & 71.99 & 74.77 & 58.33 & 79.88 & 76.00 \\
& ZEP$^\dagger$\cite{zep} & 72.34 & 77.26 & 64.58 & 83.49 & 78.94 \\
& HyperMem$^\dagger$\cite{hypermem} & \textbf{96.08} & \textbf{93.62} & \textbf{89.72} & 70.83 & 92.73 \\
\cmidrule{2-7}
& \sysname{} w/o Proactive & \underline{92.03} & 92.52 & \underline{86.59} & \underline{94.00} & \underline{92.87} \\
& \sysname{} w/o Graph Overlay & 88.48 & 92.19 & 81.25 & 93.22 & 91.39 \\
& \cellcolor{gray!12}\textbf{\sysname{} (full)} & \cellcolor{gray!12}91.84 & \cellcolor{gray!12}\underline{92.81} & \cellcolor{gray!12}80.21 & \cellcolor{gray!12}\textbf{94.89} & \cellcolor{gray!12}\textbf{92.98} \\
\bottomrule
\end{tabular}
}
\end{table*}
To overcome limitations discussed in Section \ref{sec:intro}, we further construct a benchmark, where each test instance consists of a brief user utterance (i.e. a declarative sentence containing only one memory unit) paired with a ground-truth set of units that a memory-capable model should proactively surface based on accumulated conversational history.

We adopt a bottom-up construction approach: rather than extracting memory cues post-hoc from existing dialogues, we first define atomic memory units and then weave them into naturalistic conversations through a six-step LLM-driven pipeline. 
(1) We generate a detailed user persona covering demographics, expertise, routines, and social relationships to serve as a global consistency constraint.
(2) Conditioned on the persona, we produce 60 memory units across five trigger type (i.e. temporal, entity, emotional, behavioral pattern, and multi-hop) and each represents a piece of information the user might naturally reveal.
(3) We score pairwise associations among memory units along four dimensions (entity overlap, semantic relevance, association reasonability, and conversational coherence) to form a weighted association graph. 
(4) Guided by the graph, we distribute memory units across 20 sessions (3–4 pairs each), deliberately placing highly associated pairs in \textit{different} sessions to create cross-session memory dependencies.
(5) For each session, we generate a 20-turn dialogue where memory unit information is planted naturally in user utterances.
(6) Finally, we produce 100 evaluation questions per topic, each specifying a trigger utterance and exactly three ground-truth candidate memory units that the model should proactively recall.

We repeat this pipeline for five topic domains, yielding a total of 500 test instances spanning 100 sessions and 2000 dialogue turns. Further construction details, metrics and \textbf{human validation results} are provided in Appendix~\ref{sec:appendix_proactivemembench}.

\section{Experiments}

\subsection{Experimental Setup}

\begin{table*}[t]
\centering
\caption{Comparison of \sysname{} with Memory-as-Tool methods on LongMemEval. The QA and LLM-as-Judge configurations follow \cite{longmemeval}. All experiments in this table use GPT-4o-judged systems. Baseline results are sourced from \cite{hindsight} and \cite{chrono}.}
\label{tab:longmemeval}
\resizebox{\textwidth}{!}{
\begin{tabular}{l|ccccccc}
\toprule
\textbf{Method} & \textbf{Know.-Update} & \textbf{Multi-sess.} & \textbf{Single-Asst.} & \textbf{Single-Pref.} & \textbf{Single-User} & \textbf{Temporal} & \textbf{Overall} \\
\midrule
Full Context & 78.20 & 44.30 & 94.60 & 20.00 & 81.40 & 45.10 & 60.20 \\
mem0\cite{mem0} & 79.50 & 70.70 & 46.40 & 76.70 & \underline{94.30} & 51.10 & 67.80 \\
ZEP\cite{zep} & 83.30 & 57.90 & 80.40 & 56.70 & 92.90 & 62.40 & 71.20 \\

Mastra\cite{Mastra} & 84.80 & 85.90 & 79.70 & 82.14 & 73.33 & \textbf{98.57} & 85.71 \\
Supermemory\cite{supermemory} & 81.60 & 88.50 & 71.40 & \underline{96.40} & 70.00 & 97.10 & 76.70 \\
Hindsight\cite{hindsight} & 83.60 & 84.60 & 79.70 & 94.60 & 66.70 & 95.70 & 79.70 \\

Chronos Low\cite{chrono} & \textbf{96.15} & \underline{91.73} & \textbf{100.00} & 80.00 & 94.29 & 90.23 &  \underline{92.60} \\
\midrule
\sysname{} w/o Proactive & 89.74 & \textbf{92.48} & \underline{98.21} & \textbf{96.67} & \textbf{100.00} & 96.24 & \underline{95.00} \\
\sysname{} w/o Graph Overlay & 84.42 & \underline{91.73} & \underline{98.21} & 90.00 & \textbf{100.00} & 95.45 & 93.37 \\
\cellcolor{gray!12}\textbf{\sysname{} (full)} & \cellcolor{gray!12}\underline{91.03} & \cellcolor{gray!12}\textbf{92.48} & \cellcolor{gray!12}\underline{98.21} & \cellcolor{gray!12}\textbf{96.67} & \cellcolor{gray!12}\textbf{100.00} & \cellcolor{gray!12}\underline{98.50} & \cellcolor{gray!12}\textbf{95.80} \\
\bottomrule
\end{tabular}
}
\end{table*}

\paragraph{Implementation Details.} \sysname{} is configured with navigation tools as the ReAct action space and the Proactive Reasoning Protocol injected via the system prompt. For the construction of the \memstruc{}, we use GLM-5.1\cite{glm} to build dimensions, pages, and sections. For progressive navigation and answer generation, we compare GLM-5.1, GPT-4o-mini, and GPT-4.1-mini. We limit the maximum number of tool calls and retrieval rounds to 6. 

\paragraph{Benchmarks.} We evaluate on three complementary benchmarks assessing both passive and proactive memory capabilities. \textbf{LoCoMo} \cite{locomo} is a long-conversation memory benchmark testing single-hop, multi-hop, temporal, and open-domain questions, primarily evaluating passive QA retrieval performance. \textbf{LongMemEval}\cite{longmemeval} evaluates long-term memory retrieval across extended conversational histories. \textbf{ProactiveMemBench} evaluates whether the system can spontaneously surface relevant historical information without explicit queries; we report Recall@5 and Precision judged by both LLM and human annotators.



\subsection{Main Results}

\begin{table*}[t]
\centering
\caption{Comparison of \sysname{} with Memory-as-Tool methods on ProactiveMemBench. QA and metric configurations are detailed in Appendices~\ref{subsec:proactive_generate_prompt}--\ref{subsec:llmjudgeeval}. All experiments in this table use GLM-5.1 as the backbone. All baselines are reproduced using their official implementations.}
\label{tab:proactive}
\resizebox{\textwidth}{!}{
\begin{tabular}{l|cccccc|cccccc|c}
\toprule
\multirow{2}{*}{\textbf{Method}} & \multicolumn{6}{c|}{\textbf{Recall@5}} & \multicolumn{6}{c|}{\textbf{LLM-judged Precision}} & \textbf{Human} \\
\cmidrule(lr){2-7} \cmidrule(lr){8-13} \cmidrule(lr){14-14}
& \textbf{Behav.} & \textbf{Temp.} & \textbf{Entity} & \textbf{Emot.} & \textbf{Multi-hop} & \textbf{Overall} & \textbf{Behav.} & \textbf{Temp.} & \textbf{Entity} & \textbf{Emot.} & \textbf{Multi-hop} & \textbf{Overall} & \textbf{Overall} \\
\midrule
Full Context & 53.80 & \textbf{59.09} & 62.24 & 50.24 & \underline{57.14} & \underline{56.52} & \underline{85.83} & \underline{83.27} & 87.23 & 82.71 & 84.19 & \underline{85.06} & 72.65 \\
mem0\cite{mem0} & 34.78 & 45.45 & 41.49 & 30.14 & 38.10 & 37.50 & 71.83 & 77.85 & 78.13 & 72.61 & 76.76 & 73.95 & 75.10 \\
A-Mem\cite{amem} & 23.37 & 42.42 & 30.32 & 33.33 & 47.62 & 32.75 & 68.70 & 74.10 & 75.73 & 71.65 & 77.33 & 71.67 & 75.92 \\
ZEP\cite{zep} & 47.83 & 50.01 & 53.19 & 32.84 & 42.86 & 45.98 & 75.65 & 73.64 & 81.54 & 74.00 & 79.05 & 77.12 & 80.40 \\
HyperMem\cite{hypermem} & \underline{58.51} & 48.49 & 63.47 & \underline{50.98} & 53.97 & \underline{56.52} & 78.96 & 71.45 & 84.70 & 78.35 & 80.38 & 79.13 & 85.71 \\
\midrule
\sysname{} w/o Proactive & 39.49 & 34.85 & 52.48 & 44.61 & 50.79 & 44.90 & 78.76 & 75.56 & 87.23 & 76.24 & 81.33 & 77.81 & 82.45 \\
\sysname{} w/o Graph Overlay & 50.36 & 42.42 & \underline{63.48} & 47.55 & \textbf{61.91} & 53.82 & 83.91 & 80.21 & \textbf{91.32} & \underline{87.64} & \textbf{88.19} & 83.34 & \underline{87.75} \\
\cellcolor{gray!12}\textbf{\sysname{} (full)} & \cellcolor{gray!12}\textbf{59.24} & \cellcolor{gray!12}\underline{57.58} & \cellcolor{gray!12}\textbf{71.10} & \cellcolor{gray!12}\textbf{53.43} & \cellcolor{gray!12}46.03 & \cellcolor{gray!12}\textbf{59.51} & \cellcolor{gray!12}\textbf{86.78} & \cellcolor{gray!12}\textbf{89.14} & \cellcolor{gray!12}\underline{90.96} & \cellcolor{gray!12}\textbf{88.71} & \cellcolor{gray!12}\underline{86.86} & \cellcolor{gray!12}\textbf{87.58} & \cellcolor{gray!12}\textbf{91.02} \\
\bottomrule
\end{tabular}
}
\end{table*}

\subsubsection{Passive QA Performance}

Table~\ref{tab:locomo} presents results on the LoCoMo benchmark. \sysname{} achieves 92.98 overall accuracy with GPT-4.1-mini, surpassing all baselines under the Memory-as-Tool paradigm. This result demonstrates that the Memory-as-Cognition, where multi-step navigation is interleaved with reasoning, does not sacrifice passive QA performance compared to single-shot retrieval pipelines.
Notably, \sysname{} consistently outperforms all baselines across all three QA backbone configurations (GLM: 89.83, GPT-4o-mini: 92.66, GPT-4.1-mini: 92.98), demonstrating robustness to backbones with varying reasoning capabilities and context lengths. Specifically, GLM-5.1 offers a longer context window but relatively weaker reasoning, GPT-4o-mini exhibits stronger reasoning but a shorter context window, and GPT-4.1-mini performs well in both dimensions. To illustrate the limitations of the Memory-as-Tool paradigm, we compare against its current SOTA method, HyperMem, which demands an extremely large context window during retrieval-augmented QA (nearly 30K tokens per query). Consequently, HyperMem performs only slightly below \sysname{} on long-context models, achieving 89.77 (-0.06) with GLM-5.1 and 92.73 (-0.25) with GPT-4.1-mini, but degrades substantially under low context budgets, dropping to 83.69 (-8.97) with GPT-4o-mini. This indicates that the QA model's reasoning capability fails to benefit the memory system, exposing a fundamental limitation of the Memory-as-Tool paradigm. In contrast, \sysname{} achieves consistently strong performance across all backbone configurations, confirming that its memory system design serves as effective scaffolding that compensates for both weak reasoning and limited context budgets.

\subsubsection{Long-Term Memory Retrieval}
Table~\ref{tab:longmemeval} reports results on LongMemEval. \sysname{} achieves 95.8, the highest score among all evaluated systems, outperforming baselines under Memory-as-Tool paradigm.
This confirms that \sysname{}'s multi-step navigation, which integrates the model's reasoning capability into the retrieval process, effectively handles long-term memory retrieval where information is distributed across extended interaction histories. In particular, the score on multi-session category (92.48, the highest among all methods) further substantiates this conclusion, as it requires retrieving and associating information across disparate conversation sessions. Case studies for multi-session and other categories are further provided in Appendix~\ref{appendix_navigation}.

\subsubsection{Proactive Memory}
Table~\ref{tab:proactive} presents results on ProactiveMemBench. Since conventional Memory-as-Tool methods lack proactive triggering capabilities, we augment their system prompts with proactive memory generation instructions (Appendix~\ref{subsec:proactive_generate_prompt}) while keeping retrieval methods unchanged for fair comparison. \sysname{} (full) achieves a Recall@5 of 59.51, an LLM-judged Precision of 87.58, and a human-judged Precision of 91.02, outperforming all baselines on both metrics, validating that combining the Proactive Reasoning Protocol with Cross-Dimensional Navigation yields superior proactive capabilities over the Memory-as-Tool paradigm. Removing the Proactive Reasoning Protocol (w/o Proactive) causes Recall@5 to drop from 59.51 to 44.90, confirming that the \navimemstruc{} alone is insufficient to trigger spontaneous memory exploration without proactive behavioral guidance.

We further provide a \textbf{mechanism analysis}. The proactive prompt and \navimemstruc{} form a positive feedback loop: the prompt motivates navigation initiation; the \navimemstruc{} returns results enriched with visible links and structural context; these cues naturally suggest further exploration; and the cycle reinforces itself across multiple steps. In contrast, flat memory terminates this loop after a single retrieval, as the returned top-$k$ passages contain no links, no hierarchy, and no indicators of unexplored content. The structural affordances of Pages (cross-dimensional associative links, dimensional hierarchies, summaries) continuously signal that relevant content remains accessible, sustaining multi-step exploration that flat representations cannot support.

\subsection{Ablation Studies}
\label{sec:ablation}

We conduct ablation experiments to isolate the contribution of each component. Table~\ref{tab:ablation} reports results on both LoCoMo (passive QA) and ProactiveMemBench (proactive memory triggering).

\begin{table}[t]
\centering
\caption{Ablation study results. All experiments use GLM-5.1 as the backbone.}
\label{tab:ablation}
\resizebox{\columnwidth}{!}{
\begin{tabular}{l|c|c|c}
\toprule
\multirow{2}{*}{\textbf{Configuration}} & \textbf{LoCoMo} & \multicolumn{2}{c}{\textbf{ProactiveMemBench}} \\
\cmidrule(lr){2-2} \cmidrule(lr){3-4}
& \textbf{Overall} & \textbf{Recall@5} & \textbf{LLM-judged Prec.} \\
\midrule
\cellcolor{gray!12}\textbf{\sysname{} (full)} & \cellcolor{gray!12}\textbf{89.83} & \cellcolor{gray!12}\textbf{59.51} & \cellcolor{gray!12}\textbf{87.58} \\
\midrule
w/o Proactive Protocol & \underline{89.48}{\color{blue}\scriptsize~($\downarrow$0.35)} & 44.90{\color{blue}\scriptsize~($\downarrow$14.63)} & 77.81{\color{blue}\scriptsize~($\downarrow$9.77)} \\
w/o Graph Overlay & 83.04{\color{blue}\scriptsize~($\downarrow$6.79)} & 55.32{\color{blue}\scriptsize~($\downarrow$4.19)} & \underline{83.58}{\color{blue}\scriptsize~($\downarrow$4.00)} \\
w/o Hierarchy & 83.30{\color{blue}\scriptsize~($\downarrow$6.53)} & \underline{56.59}{\color{blue}\scriptsize~($\downarrow$2.92)} & 83.36{\color{blue}\scriptsize~($\downarrow$4.22)} \\
\bottomrule
\end{tabular}
}
\end{table}

\paragraph{Proactive Reasoning Protocol.}
Removing the Proactive Reasoning Protocol has minimal impact on LoCoMo (-0.35) but devastates ProactiveMemBench (Recall@5 -14.63, LLM-judged Precision -9.77). This asymmetry confirms that the protocol specifically addresses the invocation bottleneck by activating spontaneous memory exploration, and is largely orthogonal to passive retrieval quality.

\paragraph{Navigable Memory Store.}
Removing cross-dimensional associative links (w/o Graph Overlay) degrades both LoCoMo (-6.79) and ProactiveMemBench (Recall@5 -4.19, LLM-judged Precision -4.00), confirming that associative links enable agents to follow cross-dimensional associations rather than relying on flat retrieval. Removing the dimensional hierarchy (w/o Hierarchy) causes similar degradation as hierarchical pages provide principled entry points for navigation without which the agent lacks effective starting positions for both directed search and open-ended exploration.

\paragraph{Complementary Synergy.}
The Proactive Reasoning Protocol and Navigable Memory Store contribute to different capability dimensions: the protocol enables proactive behavior while structural memory enhance retrieval quality. The full system achieves the best performance on both dimensions simultaneously, validating the Memory-as-Cognition paradigm in which iterative reasoning is interleaved with structured memory navigation.

\section{Discussion}

\subsection{Memory Access as a Spectrum}

Rather than proposing a binary dichotomy between Memory-as-Tool and Memory-as-Cognition, we argue that memory access paradigms exist along a cognitive spectrum:

\begin{center}
\textit{No Memory $\rightarrow$ Single Search $\rightarrow$ Multi-Step Navigation $\rightarrow$ Spontaneous Recall}
\end{center}

At the leftmost extreme, agents operate without any persistent memory, relying solely on the current context window. Single-shot retrieval systems occupy the next position, augmenting agents with external knowledge but limiting interaction to a single query-response cycle. \sysname{} advances the spectrum further by enabling multi-step, reasoning-driven navigation where the agent iteratively refines its exploration path. The rightmost extreme, spontaneous recall fully embedded in the agent's reasoning process without explicit tool invocation, remains a direction for future research.

This spectral view clarifies \sysname{}'s contribution: we do not claim to achieve the theoretical endpoint of memory-as-cognition, but rather demonstrate that meaningful advances along this spectrum yield substantial empirical gains. 

\subsection{On Protocol-Level Innovation}

We situate the Proactive Reasoning Protocol within a lineage of protocol-level contributions that have shaped modern LLM-based systems: Chain-of-Thought \cite{cot} for step-by-step reasoning, ReAct \cite{react} for interleaving reasoning with tool use, Self-Refine \cite{selfrefine} for iterative self-improvement, and \textbf{\sysname{}} for memory-aware proactive reasoning.

Each of these works defines \textit{how} an agent should behave in a specific cognitive dimension, and each derives its impact from the synergy between the protocol and accompanying structural innovations. In \sysname{}, the Proactive Reasoning Protocol is inseparable from the \memstruc{}: the protocol provides the cognitive drive to initiate exploration, while the Page provides the structural affordances that sustain multi-step navigation. Our ablation results (Table~\ref{tab:ablation}) empirically validate this complementary relationship, as neither component alone achieves the full system's performance.

\subsection{Scalability}
\label{subsec:scalability}
We analyze \sysname{}'s computational characteristics across three dimensions:

\paragraph{Page Construction Cost.} Initial Page construction from conversation history is $O(n)$ in the number of dialogue turns. However, incremental updates after each conversation turn operate at $O(1)$ amortized cost, as only newly extracted facts require matching and insertion. This ensures that the system remains efficient as conversation histories grow.

\paragraph{Navigation Token Overhead.} Multi-step navigation incurs additional token consumption compared to single-shot retrieval. In practice, our experiments show that agents typically perform 2--3 navigation steps per query. This overhead is modest relative to the substantial quality improvements observed, particularly on complex queries requiring enumeration, comparison, or temporal reasoning (Appendix~\ref{appendix_navigation}). Compared to HyperMem's\cite{hypermem} three-level funnel pipeline, which processes all candidate memories through multiple filtering stages, \sysname{}'s navigation is more parsimonious: the agent selectively reasons about which paths to explore at each step, avoiding exhaustive processing of irrelevant candidates and thereby achieving superior results with substantially lower token consumption.

\paragraph{Page Growth Management.} \label{sec:scalability} As conversations accumulate, the number of pages grows accordingly. We address potential page bloat through periodic archival of infrequently accessed pages and summary compression of highly detailed sections. Additionally, the hierarchical structure itself provides natural scalability. Empirically, we observe that the rate of new page creation declines by approximately 30\% over the course of extended conversations, exhibiting sub-linear growth that confirms the structure's self-regulating property. See Appendix~\ref{sec:appendix_trend} and Figure~\ref{fig:scaling_trend} for detailed analysis.

\section{Conclusion}
We present \sysname{}, a system that advances agent memory from Memory-as-Tool toward Memory-as-Cognition by integrating \memstruc{} with Proactive Reasoning Protocol, making memory access an integral part of the reasoning process. Experiments validate this paradigm shift: \sysname{} achieves state-of-the-art on passive QA benchmarks and our proposed ProactiveMemBench. Ablation studies confirm the complementary synergy between the Proactive Reasoning Protocol and Navigable Memory Store, demonstrating that advancing from passive retrieval toward reasoning-integrated navigation yields substantial improvements in agent memory capabilities.

\section{Limitations}
\sysname{} has several limitations: (1) multi-step navigation incurs additional token overhead, which may be non-trivial under strict latency or cost constraints; (2) the Proactive Reasoning Protocol relies on strong instruction-following capability of the backbone LLM, with weaker models exhibiting larger performance variance; (3) navigation quality is bounded by Page construction accuracy---poorly extracted facts or missing links degrade the entire system; and (4) with sparse early-stage conversations, the Page is under-connected and navigation offers limited advantage.

\bibliography{custom}

@misc{openhuman,
  author       = {Makel, Sena and {tinyhumansai}},
  title        = {OpenHuman: Your Personal {AI} Super Intelligence},
  year         = {2026},
  howpublished = {\url{https://github.com/tinyhumansai/openhuman}},
  note         = {Version 0.54.0. Open-source agentic assistant with memory trees, token compression, and 118+ integrations}
}

@techreport{supermemory,
  author       = {Soham Daga and Sreeram Sreedhar and Dhravya Shah},
  title        = {Supermemory: State-of-the-Art Agent Memory},
  institution  = {Supermemory Inc.},
  year         = {2026},
  type         = {Technical Report},
  url          = {https://supermemory.ai/research/},
  note         = {Accessed: 2026-05-21}
}

@misc{memr3,
      title={MemR$^3$: Memory Retrieval via Reflective Reasoning for LLM Agents}, 
      author={Xingbo Du and Loka Li and Duzhen Zhang and Le Song},
      year={2025},
      eprint={2512.20237},
      archivePrefix={arXiv},
      primaryClass={cs.AI},
      url={https://arxiv.org/abs/2512.20237}, 
}

@misc{chrono,
      title={Chronos: Temporal-Aware Conversational Agents with Structured Event Retrieval for Long-Term Memory}, 
      author={Sahil Sen and Elias Lumer and Anmol Gulati and Vamse Kumar Subbiah},
      year={2026},
      eprint={2603.16862},
      archivePrefix={arXiv},
      primaryClass={cs.CL},
      url={https://arxiv.org/abs/2603.16862}, 
}

@misc{hindsight,
      title={Hindsight is 20/20: Building Agent Memory that Retains, Recalls, and Reflects}, 
      author={Chris Latimer and Nicoló Boschi and Andrew Neeser and Chris Bartholomew and Gaurav Srivastava and Xuan Wang and Naren Ramakrishnan},
      year={2025},
      eprint={2512.12818},
      archivePrefix={arXiv},
      primaryClass={cs.CL},
      url={https://arxiv.org/abs/2512.12818}, 
}

@misc{Mastra,
  author       = {Tyler Barnes},
  title        = {Observational Memory: 95\% on {LongMemEval}},
  howpublished = {\url{https://mastra.ai/research/observational-memory}},
  month        = feb,
  year         = {2026},
  note         = {Accessed: 2026-02-22}
}

@misc{zep,
      title={Zep: A Temporal Knowledge Graph Architecture for Agent Memory}, 
      author={Preston Rasmussen and Pavlo Paliychuk and Travis Beauvais and Jack Ryan and Daniel Chalef},
      year={2025},
      eprint={2501.13956},
      archivePrefix={arXiv},
      primaryClass={cs.CL},
      url={https://arxiv.org/abs/2501.13956}, 
}

@misc{mem0,
      title={Mem0: Building Production-Ready AI Agents with Scalable Long-Term Memory}, 
      author={Prateek Chhikara and Dev Khant and Saket Aryan and Taranjeet Singh and Deshraj Yadav},
      year={2025},
      eprint={2504.19413},
      archivePrefix={arXiv},
      primaryClass={cs.CL},
      url={https://arxiv.org/abs/2504.19413}, 
}

@inproceedings{
amem,
title={A-Mem: Agentic Memory for {LLM} Agents},
author={Wujiang Xu and Zujie Liang and Kai Mei and Hang Gao and Juntao Tan and Yongfeng Zhang},
booktitle={The Thirty-ninth Annual Conference on Neural Information Processing Systems},
year={2026},
url={https://openreview.net/forum?id=FiM0M8gcct}
}

@misc{hypermem,
      title={HyperMem: Hypergraph Memory for Long-Term Conversations}, 
      author={Juwei Yue and Chuanrui Hu and Jiawei Sheng and Zuyi Zhou and Wenyuan Zhang and Tingwen Liu and Li Guo and Yafeng Deng},
      year={2026},
      eprint={2604.08256},
      archivePrefix={arXiv},
      primaryClass={cs.CL},
      url={https://arxiv.org/abs/2604.08256}, 
}

@misc{memgpt,
      title={MemGPT: Towards LLMs as Operating Systems}, 
      author={Charles Packer and Sarah Wooders and Kevin Lin and Vivian Fang and Shishir G. Patil and Ion Stoica and Joseph E. Gonzalez},
      year={2024},
      eprint={2310.08560},
      archivePrefix={arXiv},
      primaryClass={cs.AI},
      url={https://arxiv.org/abs/2310.08560}, 
}

@inproceedings{hipporag,
 author = {Guti\'{e}rrez, Bernal Jim\'{e}nez and Shu, Yiheng and Gu, Yu and Yasunaga, Michihiro and Su, Yu},
 booktitle = {Advances in Neural Information Processing Systems},
 doi = {10.52202/079017-1902},
 editor = {A. Globerson and L. Mackey and D. Belgrave and A. Fan and U. Paquet and J. Tomczak and C. Zhang},
 pages = {59532--59569},
 publisher = {Curran Associates, Inc.},
 title = {HippoRAG: Neurobiologically Inspired Long-Term Memory for Large Language Models},
 url = {https://proceedings.neurips.cc/paper_files/paper/2024/file/6ddc001d07ca4f319af96a3024f6dbd1-Paper-Conference.pdf},
 volume = {37},
 year = {2024}
}

@misc{memwalker,
      title={Walking Down the Memory Maze: Beyond Context Limit through Interactive Reading}, 
      author={Howard Chen and Ramakanth Pasunuru and Jason Weston and Asli Celikyilmaz},
      year={2023},
      eprint={2310.05029},
      archivePrefix={arXiv},
      primaryClass={cs.CL},
      url={https://arxiv.org/abs/2310.05029}, 
}

@misc{raptor,
      title={RAPTOR: Recursive Abstractive Processing for Tree-Organized Retrieval}, 
      author={Parth Sarthi and Salman Abdullah and Aditi Tuli and Shubh Khanna and Anna Goldie and Christopher D. Manning},
      year={2024},
      eprint={2401.18059},
      archivePrefix={arXiv},
      primaryClass={cs.CL},
      url={https://arxiv.org/abs/2401.18059}, 
}

@inproceedings{ircot,
    title = "Interleaving Retrieval with Chain-of-Thought Reasoning for Knowledge-Intensive Multi-Step Questions",
    author = "Trivedi, Harsh  and
      Balasubramanian, Niranjan  and
      Khot, Tushar  and
      Sabharwal, Ashish",
    editor = "Rogers, Anna  and
      Boyd-Graber, Jordan  and
      Okazaki, Naoaki",
    booktitle = "Proceedings of the 61st Annual Meeting of the Association for Computational Linguistics (Volume 1: Long Papers)",
    month = jul,
    year = "2023",
    address = "Toronto, Canada",
    publisher = "Association for Computational Linguistics",
    url = "https://aclanthology.org/2023.acl-long.557/",
    doi = "10.18653/v1/2023.acl-long.557",
    pages = "10014--10037"
}

@misc{generativeagents,
      title={Generative Agents: Interactive Simulacra of Human Behavior}, 
      author={Joon Sung Park and Joseph C. O'Brien and Carrie J. Cai and Meredith Ringel Morris and Percy Liang and Michael S. Bernstein},
      year={2023},
      eprint={2304.03442},
      archivePrefix={arXiv},
      primaryClass={cs.HC},
      url={https://arxiv.org/abs/2304.03442}, 
}

@misc{proactivedialogue,
      title={A Survey on Proactive Dialogue Systems: Problems, Methods, and Prospects},
      author={Yang Deng and Wenqiang Lei and Wai Lam and Tat-Seng Chua},
      year={2023},
      eprint={2305.02750},
      archivePrefix={arXiv},
      primaryClass={cs.CL},
      url={https://arxiv.org/abs/2305.02750},
}

@misc{langmem,
      title={LangMem},
      author={LangChain},
      year={2026},
      url={https://github.com/langchain-ai/langmem},
      note={Software library for long-term memory in LLM agents},
}

@inproceedings{adaptiverag,
      title={Adaptive-RAG: Learning to Adapt Retrieval-Augmented Large Language Models through Question Complexity},
      author={Soyeong Jeong and Jinheon Baek and Sukmin Cho and Sung Ju Hwang and Jong C. Park},
      booktitle={Proceedings of the 2024 Conference of the North American Chapter of the Association for Computational Linguistics},
      year={2024},
      url={https://arxiv.org/abs/2403.14403},
}

@misc{crag,
      title={Corrective Retrieval Augmented Generation},
      author={Shi-Qi Yan and Jia-Chen Gu and Yun Zhu and Zhen-Hua Ling},
      year={2024},
      eprint={2401.15884},
      archivePrefix={arXiv},
      primaryClass={cs.CL},
      url={https://arxiv.org/abs/2401.15884},
}

@misc{speculativerag,
      title={Speculative RAG: Enhancing Retrieval Augmented Generation through Drafting},
      author={Zilong Wang and Zifeng Wang and Long Le and Huaixiu Steven Zheng and Swaroop Mishra and Vincent Perot and Yuwei Zhang and Anush Mattapalli and Ankur Taly and Jingbo Shang and Chen-Yu Lee and Tomas Pfister},
      year={2024},
      eprint={2407.08223},
      archivePrefix={arXiv},
      primaryClass={cs.CL},
      url={https://arxiv.org/abs/2407.08223},
}

@inproceedings{
react,
title={ReAct: Synergizing Reasoning and Acting in Language Models},
author={Shunyu Yao and Jeffrey Zhao and Dian Yu and Nan Du and Izhak Shafran and Karthik R Narasimhan and Yuan Cao},
booktitle={The Eleventh International Conference on Learning Representations },
year={2023},
url={https://arxiv.org/abs/2210.03629}, 
}

@inproceedings{
longmemeval,
title={LongMemEval: Benchmarking Chat Assistants on Long-Term Interactive Memory},
author={Di Wu and Hongwei Wang and Wenhao Yu and Yuwei Zhang and Kai-Wei Chang and Dong Yu},
booktitle={The Thirteenth International Conference on Learning Representations},
year={2025},
url={https://arxiv.org/abs/2410.10813}, 
}

@inproceedings{locomo,
    title = "Evaluating Very Long-Term Conversational Memory of {LLM} Agents",
    author = "Maharana, Adyasha  and
      Lee, Dong-Ho  and
      Tulyakov, Sergey  and
      Bansal, Mohit  and
      Barbieri, Francesco  and
      Fang, Yuwei",
    editor = "Ku, Lun-Wei  and
      Martins, Andre  and
      Srikumar, Vivek",
    booktitle = "Proceedings of the 62nd Annual Meeting of the Association for Computational Linguistics (Volume 1: Long Papers)",
    month = aug,
    year = "2024",
    address = "Bangkok, Thailand",
    publisher = "Association for Computational Linguistics",
    url = "https://aclanthology.org/2024.acl-long.747/",
    doi = "10.18653/v1/2024.acl-long.747",
    pages = "13851--13870",
    
}

@misc{glm,
      title={GLM-5: from Vibe Coding to Agentic Engineering}, 
      author={GLM-5-Team and : and Aohan Zeng and Xin Lv and Zhenyu Hou and Zhengxiao Du and Qinkai Zheng and Bin Chen and Da Yin and Chendi Ge and Chenghua Huang and Chengxing Xie and Chenzheng Zhu and Congfeng Yin and Cunxiang Wang and Gengzheng Pan and Hao Zeng and Haoke Zhang and Haoran Wang and Huilong Chen and Jiajie Zhang and Jian Jiao and Jiaqi Guo and Jingsen Wang and Jingzhao Du and Jinzhu Wu and Kedong Wang and Lei Li and Lin Fan and Lucen Zhong and Mingdao Liu and Mingming Zhao and Pengfan Du and Qian Dong and Rui Lu and Shuang-Li and Shulin Cao and Song Liu and Ting Jiang and Xiaodong Chen and Xiaohan Zhang and Xuancheng Huang and Xuezhen Dong and Yabo Xu and Yao Wei and Yifan An and Yilin Niu and Yitong Zhu and Yuanhao Wen and Yukuo Cen and Yushi Bai and Zhongpei Qiao and Zihan Wang and Zikang Wang and Zilin Zhu and Ziqiang Liu and Zixuan Li and Bojie Wang and Bosi Wen and Can Huang and Changpeng Cai and Chao Yu and Chen Li and Chengwei Hu and Chenhui Zhang and Dan Zhang and Daoyan Lin and Dayong Yang and Di Wang and Ding Ai and Erle Zhu and Fangzhou Yi and Feiyu Chen and Guohong Wen and Hailong Sun and Haisha Zhao and Haiyi Hu and Hanchen Zhang and Hanrui Liu and Hanyu Zhang and Hao Peng and Hao Tai and Haobo Zhang and He Liu and Hongwei Wang and Hongxi Yan and Hongyu Ge and Huan Liu and Huanpeng Chu and Jia'ni Zhao and Jiachen Wang and Jiajing Zhao and Jiamin Ren and Jiapeng Wang and Jiaxin Zhang and Jiayi Gui and Jiayue Zhao and Jijie Li and Jing An and Jing Li and Jingwei Yuan and Jinhua Du and Jinxin Liu and Junkai Zhi and Junwen Duan and Kaiyue Zhou and Kangjian Wei and Ke Wang and Keyun Luo and Laiqiang Zhang and Leigang Sha and Liang Xu and Lindong Wu and Lintao Ding and Lu Chen and Minghao Li and Nianyi Lin and Pan Ta and Qiang Zou and Rongjun Song and Ruiqi Yang and Shangqing Tu and Shangtong Yang and Shaoxiang Wu and Shengyan Zhang and Shijie Li and Shuang Li and Shuyi Fan and Wei Qin and Wei Tian and Weining Zhang and Wenbo Yu and Wenjie Liang and Xiang Kuang and Xiangmeng Cheng and Xiangyang Li and Xiaoquan Yan and Xiaowei Hu and Xiaoying Ling and Xing Fan and Xingye Xia and Xinyuan Zhang and Xinze Zhang and Xirui Pan and Xu Zou and Xunkai Zhang and Yadi Liu and Yandong Wu and Yanfu Li and Yidong Wang and Yifan Zhu and Yijun Tan and Yilin Zhou and Yiming Pan and Ying Zhang and Yinpei Su and Yipeng Geng and Yong Yan and Yonglin Tan and Yuean Bi and Yuhan Shen and Yuhao Yang and Yujiang Li and Yunan Liu and Yunqing Wang and Yuntao Li and Yurong Wu and Yutao Zhang and Yuxi Duan and Yuxuan Zhang and Zezhen Liu and Zhengtao Jiang and Zhenhe Yan and Zheyu Zhang and Zhixiang Wei and Zhuo Chen and Zhuoer Feng and Zijun Yao and Ziwei Chai and Ziyuan Wang and Zuzhou Zhang and Bin Xu and Minlie Huang and Hongning Wang and Juanzi Li and Yuxiao Dong and Jie Tang},
      year={2026},
      eprint={2602.15763},
      archivePrefix={arXiv},
      primaryClass={cs.LG},
      url={https://arxiv.org/abs/2602.15763}, 
}

@inproceedings{cot,
 author = {Wei, Jason and Wang, Xuezhi and Schuurmans, Dale and Bosma, Maarten and ichter, brian and Xia, Fei and Chi, Ed and Le, Quoc V and Zhou, Denny},
 booktitle = {Advances in Neural Information Processing Systems},
 editor = {S. Koyejo and S. Mohamed and A. Agarwal and D. Belgrave and K. Cho and A. Oh},
 pages = {24824--24837},
 publisher = {Curran Associates, Inc.},
 title = {Chain-of-Thought Prompting Elicits Reasoning in Large Language Models},
 url = {https://proceedings.neurips.cc/paper_files/paper/2022/file/9d5609613524ecf4f15af0f7b31abca4-Paper-Conference.pdf},
 volume = {35},
 year = {2022}
}

@inproceedings{selfrefine,
 author = {Madaan, Aman and Tandon, Niket and Gupta, Prakhar and Hallinan, Skyler and Gao, Luyu and Wiegreffe, Sarah and Alon, Uri and Dziri, Nouha and Prabhumoye, Shrimai and Yang, Yiming and Gupta, Shashank and Majumder, Bodhisattwa Prasad and Hermann, Katherine and Welleck, Sean and Yazdanbakhsh, Amir and Clark, Peter},
 booktitle = {Advances in Neural Information Processing Systems},
 editor = {A. Oh and T. Naumann and A. Globerson and K. Saenko and M. Hardt and S. Levine},
 pages = {46534--46594},
 publisher = {Curran Associates, Inc.},
 title = {Self-Refine: Iterative Refinement with Self-Feedback},
 url = {https://proceedings.neurips.cc/paper_files/paper/2023/file/91edff07232fb1b55a505a9e9f6c0ff3-Paper-Conference.pdf},
 volume = {36},
 year = {2023}
}

\appendix
\clearpage
\section{Memory-Aware Reasoning Protocol}
\label{sec:appendix}
\subsection{Protocol Definition}
The following is the complete Proactive Reasoning Protocol provided to the agent as a system prompt. This protocol defines the behavioral norms for memory usage, including when and how to initiate memory navigation.

\begin{tcolorbox}[
  colback=gray!5,
  colframe=gray!60,
  title={\textbf{Proactive Reasoning Protocol (System Prompt)}},
  fonttitle=\small,
  boxrule=0.5pt,
  arc=2pt,
  left=6pt, right=6pt, top=4pt, bottom=4pt
]
{\small\ttfamily
\textbf{\#\# Your Memory System}\\[4pt]
You have access to a \memstruc{} about the user, organized by dimensions (personal preferences, work, relationships, events, etc.).\\[6pt]
\textbf{\#\#\# When to use memory}\\[4pt]
\textbf{Proactive association} (most important):\\
- When the user mentions people, places, topics, or situations that may connect to stored knowledge → navigate your memory to check for relevant context\\
- When you sense the conversation could benefit from historical context → browse relevant dimensions\\
- When the user's statement might contradict or update stored information → verify against memory\\[4pt]
\textbf{Reactive response} (traditional):\\
- When the user explicitly asks about past information → search or navigate\\[6pt]
\textbf{\#\#\# How to use memory}\\[4pt]
- \textbf{Navigate progressively}: Don't stop at the first result. If a page has relevant links, follow them. Build a complete picture.\\
- \textbf{Integrate naturally}: Use recalled information as if you naturally remember it. Don't say "according to my records."\\
- \textbf{Be selective}: Not everything you find needs to be shared. Consider whether the information is relevant, timely, and appropriate.\\
- \textbf{Think before acting}: Before navigating, briefly reason about which dimension/topic is most likely relevant.
}
\end{tcolorbox}

\subsection{Protocol as Cognitive Activation}

The Proactive Reasoning Protocol activates the cognitive potential of the \navimemstruc{}. We provide three arguments for this claim:

First, protocol-level design has partial precedent as a validated contribution. ReAct \cite{react} defines the Think$\rightarrow$Act$\rightarrow$Obs format in prompts and represents a landmark contribution in agent reasoning. Proactive Reasoning Protocol similarly defines the agent's behavioral pattern for memory usage.

Second, the protocol and structure exhibit complementary synergy that neither achieves alone. A proactive instruction paired with flat memory yields limited improvement because returned passages offer no structural cues for continued exploration. A \navimemstruc{} without Proactive Reasoning Protocol remains underutilized because the agent defaults to passive, query-triggered behavior. Only the combination produces the intended effect: the protocol motivates navigation initiation, while the \navimemstruc{} sustains multi-step exploration through its inherent navigational affordances.

Third, this synergy is empirically verifiable through ablation experiments. Removing only Proactive Reasoning Protocol (\navimemstruc{} retained) substantially degrades proactive association performance while preserving passive QA capability. Removing only the \navimemstruc{} (proactive prompt retained with flat memory) limits the depth of exploration regardless of the agent's motivation to explore. \sysname{} adopts both the proactive prompt and the \navimemstruc{}, yielding significant synergistic gains.

\section{ProactiveMemBench}
\label{sec:appendix_proactivemembench}
\subsection{Motivation}
Existing memory benchmarks (e.g., LoCoMo~\cite{locomo}) primarily evaluate reactive memory retrieval, which is the ability to answer explicit questions about past conversations. However, in real-world conversational AI, a more valuable yet understudied capability is \textbf{proactive memory association}: when a user mentions concept $A$ in the current turn, the model should spontaneously recall and surface related concepts $B$, $C$, $D$ from prior conversations which the user has not explicitly asked about but would benefit from being reminded of.

To systematically evaluate this ability, we construct \textbf{ProactiveMemBench}, a benchmark where each test instance consists of a brief user utterance (i.e. a declarative sentence containing only one memory unit) paired with a ground-truth set of memory units that a memory-capable model should proactively surface based on accumulated conversational history.

\subsection{Construction}

We identify three core design principles:

\textbf{Bottom-up construction with persona consistency.} Rather than generating dialogues first and deriving questions post-hoc, we adopt a bottom-up approach: we first define atomic memory units and then weave them into coherent dialogues. A fixed user persona constrains all generation steps to ensure global consistency.

\textbf{Multi-type memory associations.} We define five distinct types of proactive memory triggers that mirror how human memory works, which are illustrated in Table ~\ref{tab:trigger_types}.

\textbf{Graph-structured association labeling.} Memory Units are not treated as independent ones; instead, we explicitly model their pairwise associations, forming a graph that enables controlled placement of cross-session memory cues.

The benchmark is constructed through a six-step LLM-driven pipeline, using Qwen3.5-397B-A17B as the generation backbone. Each step's output is conditioned on all preceding outputs to maintain coherence.

\textbf{Step 1: User Persona Generation} \\
We first manually generate a detailed, internally consistent user persona for each topic domain. The persona covers eight dimensions: basic demographics, domain expertise, family relationships, social circle, daily routines, emotion-behavior mappings, life milestones, and environmental preferences. This persona serves as a global constraint injected into all subsequent generation prompts, preventing contradictions across steps.

\textbf{Step 2: Memory Unit Generation}

Based on each persona, we generate 60 memory units per topic (12 per trigger type), following a strict sequential order: temporal → entity → emotional → behavioral → multi-hop. Each subsequent type receives the persona plus all previously generated pairs as context, ensuring no factual contradictions between pairs, entities and events consistent with the persona, and Multi-hop chains grounded in verified single-hop relationships.

The following prompt is used to generate memory units. We show the \emph{temporal trigger} variant as an example; the other four types (entity, emotional, behavioral pattern, multi-hop) follow the same template with type-specific instructions. Crucially, this prompt is \textbf{preceded by the full user persona} and \textbf{all previously generated memory units} to prevent contradictions.

\begin{tcolorbox}[colback=blue!3, colframe=blue!40,breakable,   fonttitle=\small,title=\textbf{Memory Unit Generation Prompt (Temporal Trigger Type)}]
\small
\texttt{You are a data construction expert. Please generate memory units strictly based on the given user persona. Output a JSON array directly.}

\medskip
\texttt{[Global Constraint: User Persona]}\\
\textit{\{Full persona JSON is injected here\}}

\medskip
\texttt{[Previously Generated Memory Units]}\\
\textit{\{All memory units from earlier types are appended here\}}

\medskip
\texttt{[Task]}\\
Based on the user persona, generate \textbf{12} ``temporal trigger'' memory units.\\
Temporal trigger = a music-related activity the user performs at a specific/periodic time.

\medskip
\textbf{Requirements:}\\
(1) Must be consistent with the persona's \texttt{daily\_habits}; extend with more details.\\
(2) Diversify time patterns: daily, specific weekday, monthly, seasonal, holidays.\\
(3) Music activities must be specific: name composers, pieces, and practice content.

\medskip
\textbf{Output format (JSON array):}\\
\texttt{[\{"id": "time\_01", "type": "temporal trigger", "time\_pattern": "every Monday evening at 8pm", "music\_entity": "practice Bach WTC", "natural\_expression": "I practice Bach's WTC every Monday evening.", "involved\_entities": ["Bach","WTC","piano"]\}]}
\end{tcolorbox}

\paragraph{Type-specific variations.}
\begin{itemize}[nosep]
    \item \textbf{Entity trigger}: Generate 12 memory units binding a music entity to people, places, or evaluations (e.g., ``My grandmother loves Beethoven's Moonlight Sonata'').
    \item \textbf{Emotional trigger}: Generate 12 memory units linking emotions to music behaviors, covering both directions (emotion$\to$music and music$\to$emotion).
    \item \textbf{Behavioral pattern trigger}: Generate 12 memory units capturing habitual co-occurrences of music activities and companion behaviors (e.g., ``I always brew green tea before practicing'').
    \item \textbf{Multi-hop trigger}: Generate 12 inference chains (3--4 nodes) by connecting entities from previously generated memory units, requiring $\geq$2 reasoning hops.
\end{itemize}

\textbf{Step 3: Association Scoring}

\begin{table*}[ht]
\centering
\small
\begin{tabular}{ll}
\toprule
\textbf{Dimension} & \textbf{Description} \\
\midrule
Entity overlap & Shared named entities \\
Semantic relevance & Topical proximity \\
Association reasonability & Whether recalling $B$ upon hearing $A$ feels natural \\
Conversational coherence & Accompanying habits when the user performs certain actions \\
Multi-hop Association & Whether transitioning from $A$ to $B$ in dialogue is smooth \\
\bottomrule
\end{tabular}
\caption{Evaluation Dimensions in Association Scoring.}
\label{tab:dimensions}
\end{table*}
To model the relatedness structure among memory units, we evaluate each pair of them along four dimensions showed in Table ~\ref{tab:dimensions}. 
This prompt evaluates pairwise associations between memory units. Memory units are processed in batches of 10.

\begin{tcolorbox}[colback=blue!3, colframe=blue!40,breakable,   fonttitle=\small,title=\textbf{Association Scoring Prompt}]
\small
\texttt{You are a semantic association evaluation expert. Please assess the association strength between given memory unit combinations. Output a JSON array directly.}

\medskip
Please evaluate the following \textit{N} memory unit combinations.

For each combination, score along 4 dimensions (1--5 scale):\\
(1) \textbf{Entity overlap}: Do the two memory units share named entities?\\
(2) \textbf{Semantic relevance}: How close are they topically?\\
(3) \textbf{Association reasonability} (highest weight): Would recalling B upon hearing A feel natural?\\
(4) \textbf{Conversational coherence}: Would transitioning from A to B in dialogue be smooth?

\medskip
\textbf{Scoring rubric:}\\
5 = Strong association (mentioning A should almost certainly trigger B)\\
4 = Moderate-strong (associating B from A is natural)\\
3 = Moderate (some connection but not obligatory)\\
2 = Weak (indirect link; proactive mention may feel abrupt)\\
1 = No association (should not proactively recall)


\medskip
\textbf{Output format:}\\
\texttt{[\{"pair\_a\_id": "time\_01", "pair\_b\_id": "entity\_03", "shared\_entities": ["piano"], "scores": \{"entity\_overlap": 2, "semantic\_relevance": 3, "association\_reasonability": 3, "conversation\_coherence": 4\},  "association\_path": "Both involve piano; Bach and Beethoven are classical composers."\}]}

\medskip
\textbf{Memory units to evaluate:}\\
\textit{\{Batch of 10 memory unit combinations in JSON\}}
\end{tcolorbox}

\textbf{Step 4: Session Grouping via Graph-Aware Allocation}

We distribute the 60 memory units across 20 dialogue sessions (3–4 units each), applying a key constraint derived from the association graph: highly associated pairs (overall score $\geq 3.5$) are deliberately assigned to different sessions, which creates cross-session memory dependencies.

\textbf{Step 5: Dialogue Generation}

For each session, we generate 20 dialogue turns. 

This prompt generates a 20-turn session dialogue. The full user persona is provided as a global constraint.

\begin{tcolorbox}[colback=blue!3, colframe=blue!40,breakable,   fonttitle=\small,title=\textbf{Dialogue Generation Prompt}]
\small
\texttt{You are a dialogue generation expert. Generate natural, realistic multi-turn conversations on music topics based on the given session information and user persona.}

\medskip
\texttt{[User Persona]}\\
\textit{\{Full persona JSON\}}

\medskip
\textbf{Rules:}\\
(1) Dialogue content must not contradict the user persona.\\
(2) People, places, and experiences must be consistent with the persona.\\
(3) The user's language should sound like a real young person.

\medskip
Generate 20 turns for Session \textit{id} (user 10 turns + assistant 10 turns, alternating).

\medskip
\textbf{Session info:}\\
- Time: \textit{\{timestamp\_range\}}\\
- Theme: \textit{\{session\_theme\}}\\
- Conversation seed: \textit{\{conversation\_seed\}}

\medskip
\textbf{Memory units to plant:}\\
\textit{\{JSON of assigned memory units with natural\_expression and involved\_entities\}}

\medskip
\textbf{Requirements:}\\
(1) Total 20 alternating turns.\\
(2) Each memory unit must be naturally mentioned at least once.\\
(3) Plantings should be \emph{dispersed} (not clustered in consecutive turns).\\
(4) Use colloquial, authentic language.\\
(5) Memory unit information is revealed by the \textbf{user} (not elicited by assistant).\\
(6) Each turn must carry an ISO 8601 timestamp.

\medskip
\textbf{Output format:}\\
\texttt{\{"session\_id": 1, "turns": [\{"turn\_id": 1, "role": "user", "content": "Hey, it's Monday...", "timestamp": "2024-03-01T20:15:00", "planted\_units": ["time\_01"], "mentioned\_entities": ["piano","Monday"]\}], "unit\_coverage": \{"time\_01": \{"planted": true, "turns": [1,3]\}\}\}}
\end{tcolorbox}

\textbf{Step 6: Proactive Question Generation}

Finally, we generate 100 evaluation questions per topic (20 per trigger type), each following the structure:
$q = (trigger\_utterance, candidate\_set, difficulty)$, 
where $trigger\_utterance$ refers to a brief user message containing only one or two memory units; $candidate\_set$ refers to exactly 3 memory units the model should recall; and $Difficulty$ refers to the difficulty of the question.

This prompt generates evaluation questions. It is run once per trigger type, producing 20 questions each (100 total).

\begin{tcolorbox}[colback=blue!3, colframe=blue!40,breakable, title=\textbf{Proactive Question Generation Prompt},  fonttitle=\small]
\small
\texttt{You are a benchmark design expert. You need to generate evaluation questions for the ``proactive memory association'' capability.}

\medskip
\texttt{[User Persona]}\\
\textit{\{Full persona JSON\}}

\medskip
\textbf{Background:}\\
``Proactive memory association'' means: when a user mentions concept A in the current turn, the model should---based on memorized conversation history---proactively recall related concepts B, C, D that the user is \emph{not} currently mentioning but has previously shared.

\medskip
Generate \textit{count} proactive evaluation questions based on ``\textit{trigger\_type}''.

\medskip
\textbf{Memory units of this type:}\\
\textit{\{JSON of type-specific memory units\}}

\medskip
\textbf{Relevant cross-type associations:}\\
\textit{\{JSON of high-association memory units involving this type\}}

\medskip
\textbf{Available entity pool:}\\
\textit{\{All entities extracted from memory units\}}

\medskip
\textbf{Core design logic:}\\
The user says A in the current turn; the model should recall B, C, D from history---things the user is \emph{not} saying now but mentioned before.

\medskip
\textbf{Anti-pattern} (DO NOT do this):\\
- User says ``I listen to BBC Radio 3 every morning'' $\to$ \texttt{candidate\_set} = [``BBC Radio 3'']\\
- \textcolor{red}{Wrong!} The user already said these; the model doesn't need to ``recall'' them.

\medskip
    \textbf{Correct pattern:}\\
- User says ``I heard a Debussy piece on the radio this morning''\\
- \texttt{candidate\_set}: [``Moonlight (piece user is practicing)'', ``Teacher Zhang (guides tone)'', ``Wednesday practice day (also plays Debussy)'']\\
- These are things the user did \emph{not} say, but the model should proactively surface from prior sessions.

\medskip
\textbf{Question structure:}\\
(1) \texttt{question}: A short user utterance containing only 1--2 trigger memory units.\\
(2) \texttt{trigger\_memory\_unit}: The core trigger word in the question.\\
(3) \texttt{candidate\_set}: 3 memory units the model should recall from history (MUST be absent from the question).\\
(4) \texttt{difficulty}: easy (1-hop) / medium (2-hop) / hard (cross-session multi-hop).\\
(5) \texttt{reasoning}: Full association chain.

\medskip
\textbf{Difficulty distribution:} easy: 6, medium: 8, hard: 6.

\medskip
\textbf{Output format:}\\
\texttt{[\{"id": "q\_001", "trigger\_type": "temporal trigger", "question": "short user utterance", "trigger\_memory\_unit": "trigger word", "candidate\_set": [\{"memory\_unit": "A", "reason": "path..."\}, \{"memory\_unit": "B", "reason": "path..."\}, \{"memory\_unit": "C", "reason": "path..."\}], "difficulty": "easy/medium/hard", "source\_units": ["time\_01","entity\_03"], "reasoning": "full explanation"\}]}
\end{tcolorbox}

ProactiveMemBench evaluates five types of proactive memory triggers. Table~\ref{tab:trigger_types} summarizes the trigger type taxonomy used in the benchmark.

\begin{table*}[ht]
\centering
\small
\begin{tabular}{ll}
\toprule
\textbf{Trigger Type} & \textbf{Description} \\
\midrule
Temporal Association & Things the user does at specific times/cycles \\
Entity Association & User's comparative evaluations, associations, or bindings with entities/people \\
Emotional Association & Bindings between specific emotions and user behaviors \\
Behavioral Pattern Association & Accompanying habits when the user performs certain actions \\
Multi-hop Association & Unit chains requiring 2--3 hops of reasoning through intermediate entities \\
\bottomrule
\end{tabular}
\caption{Trigger type taxonomy in ProactiveMemBench.}
\label{tab:trigger_types}
\end{table*}

\subsection{Generate Prompt}
\label{subsec:proactive_generate_prompt}
The following prompt is used to instruct the model to retrieve relevant triggered memory units from the memory store given a user message.

\begin{tcolorbox}[
  colback=blue!5,
  colframe=blue!40,
  breakable,
  title={\textbf{Generate Prompt}},
  fonttitle=\small,
  boxrule=0.5pt,
  arc=2pt,
  left=6pt, right=6pt, top=4pt, bottom=4pt
]
{\small\ttfamily
You are a memory retrieval assistant. Given a user message and relevant memory fragments, identify the most relevant triggered memory units based on the memory content. You only need to output the list of retrieved memory units in the specified JSON format, with no extra text.\\[6pt]
---\\
\textbf{Examples of triggered memory units}:\\
\{few\_shot\}\\[6pt]
---\\[4pt]
\textbf{Now please complete the following task}:\\[4pt]
\textbf{Trigger type}: \{trigger\_type\} (\{desc\})\\[4pt]
\textbf{User message}: ``\{question\}''\\[4pt]
Please retrieve from the memory store and find the most relevant \textbf{up to 5} triggered memories for this user message. Each memory should include:\\
- memory\_unit: memory unit name\\
- reason: association path explanation\\[4pt]
\textbf{Please strictly output in the following JSON format, with no extra text}:\\
{[}\\
\hspace*{1em}\{``memory\_unit'': ``...'', ``reason'': ``\textbf{Association path}: ...''\},\\
\hspace*{1em}...\\
{]}\\[6pt]
---\\
The following are relevant memory fragments retrieved from the memory store (A-MEM) for your reference:\\
\{mem\_context\}\\[4pt]
Based on the above memories, output up to 5 triggered memories (in JSON format).
}
\end{tcolorbox}

\noindent\textbf{Variable Descriptions:}
\begin{itemize}
    \item \texttt{\{few\_shot\}}: Few-shot examples of triggered memory units, providing format and content references for the model.
    \item \texttt{\{trigger\_type\}}: The type of memory trigger, e.g., temporal, entity-related, emotional, behavioral pattern, or multi-hop.
    \item \texttt{\{desc\}}: A brief description of the trigger type explaining what kind of memory association it refers to.
    \item \texttt{\{question\}}: The user's input message that serves as the query for memory retrieval.
    \item \texttt{\{mem\_context\}}: Relevant memory fragments retrieved from the A-MEM memory store, provided as contextual reference.
\end{itemize}

\noindent The following box shows a representative few-shot example for each trigger type, illustrating the content of the \texttt{\{few\_shot\}} variable:

\begin{tcolorbox}[
  colback=blue!3,
  colframe=blue!30,
  breakable,
  title={\textbf{Few-shot Examples by Trigger Type}},
  fonttitle=\small,
  boxrule=0.5pt,
  arc=2pt,
  left=6pt, right=6pt, top=4pt, bottom=4pt
]
{\small\ttfamily
\textbf{1. Temporal Association}\\
User message: ``It's almost New Year, anything I should do?''\\
\textbf{Triggered memories}:\\
{[}\\
\hspace*{1em}\{``memory\_unit'': ``Promised kid a year-end trip'',\\
\hspace*{1.5em}``reason'': ``\textbf{Association path}: user says it's almost New Year $\rightarrow$ find promise of year-end trip with kid''\}\\
{]}\\[8pt]
\textbf{2. Entity Association}\\
User message: ``How is Tim doing lately?''\\
\textbf{Triggered memories}:\\
{[}\\
\hspace*{1em}\{``memory\_unit'': ``AI startup project progress'',\\
\hspace*{1.5em}``reason'': ``\textbf{Association path}: user mentions Tim $\rightarrow$ find co-founded startup in Hangzhou $\rightarrow$ find AI startup project progress''\}\\
{]}\\[8pt]
\textbf{3. Emotional Association}\\
User message: ``Feeling terrible, don't want to do anything.''\\
\textbf{Triggered memories}:\\
{[}\\
\hspace*{1em}\{``memory\_unit'': ``User likes running to relieve stress'',\\
\hspace*{1.5em}``reason'': ``\textbf{Association path}: user expresses negative emotion $\rightarrow$ scan relaxation/hobby pages $\rightarrow$ find user likes running to relieve stress''\}\\
{]}\\[8pt]
\textbf{4. Behavioral Pattern Association}\\
User message: ``Just improvised on the piano and recorded it.''\\
\textbf{Triggered memories}:\\
{[}\\
\hspace*{1em}\{``memory\_unit'': ``Architecture proposal ambient music material'',\\
\hspace*{1.5em}``reason'': ``\textbf{Association path}: user mentions improvising piano and recording $\rightarrow$ find user previously used recordings as ambient music material for architecture proposals $\rightarrow$ suggest adding to material library''\}\\
{]}\\[8pt]
\textbf{5. Multi-hop Association}\\
User message: ``Where is Joe's hometown?''\\
\textbf{Triggered memories}:\\
{[}\\
\hspace*{1em}\{``memory\_unit'': ``Yantai'',\\
\hspace*{1.5em}``reason'': ``\textbf{Association path}: user asks about Joe's hometown $\rightarrow$ find Joe once gifted hometown specialty Yantai apples $\rightarrow$ answer is Yantai''\}\\
{]}
}
\end{tcolorbox}

\subsection{Evaluation Metrics: Recall@k}

The Recall@k metric evaluates whether the model's retrieved memory units semantically match the ground-truth candidate memory units. An LLM judge determines whether each candidate memory unit has a corresponding match in the retrieval results.

\begin{tcolorbox}[
  colback=green!5,
  colframe=green!40,
  breakable,
  title={\textbf{Recall@k Judge Prompt -- System}},
  fonttitle=\small,
  boxrule=0.5pt,
  arc=2pt,
  left=6pt, right=6pt, top=4pt, bottom=4pt
]
{\small\ttfamily
You are an evaluation expert responsible for determining whether the content retrieved by the model from the memory store matches the ground-truth answers. Your task is: for each memory unit in the ground-truth answers, determine whether the model's retrieval results contain semantically identical or highly similar content. Output must be in JSON format only, with no extra text.
}
\end{tcolorbox}

\begin{tcolorbox}[
  colback=green!5,
  colframe=green!40,
  breakable,
  title={\textbf{Recall@k Judge Prompt -- User}},
  fonttitle=\small,
  boxrule=0.5pt,
  arc=2pt,
  left=6pt, right=6pt, top=4pt, bottom=4pt
]
{\small\ttfamily
\textbf{User message}: \{question\}\\[6pt]
\textbf{Memory unit list from ground-truth (candidate\_set)}:\\
\{candidate\_units\}\\[6pt]
\textbf{Memory unit list from model retrieval results}:\\
\{retrieved\_units\}\\[6pt]
\textbf{Please determine}: for memory each unit in the ground-truth, whether there is a semantically identical or highly similar match in the model's retrieval results.\\
Note: Exact wording does not need to match, but the core memory unit must be the same.\\[4pt]
\textbf{Please output in the following JSON format}:\\
\{\\
\hspace*{1em}``matches'': {[}\\
\hspace*{2em}\{\\
\hspace*{3em}``candidate\_unit'': ``memory unit from ground-truth'',\\
\hspace*{3em}``matched'': true or false,\\
\hspace*{3em}``matched\_retrieved\_unit'': ``if matched, fill in the corresponding memory unit, otherwise null''\\
\hspace*{2em}\}\\
\hspace*{1em}{]}\\
\}
}
\end{tcolorbox}

\noindent\textbf{Variable Descriptions:}
\begin{itemize}
    \item \texttt{\{question\}}: The user's input message serving as the retrieval query.
    \item \texttt{\{candidate\_units\}}: The list of ground-truth memory units (JSON-formatted) that the model is expected to retrieve.
    \item \texttt{\{retrieved\_units\}}: The list of memory units (JSON-formatted) actually retrieved by the model.
\end{itemize}

\subsection{Evaluation Metrics: Precision}
\label{subsec:llmjudgeeval}
The Precision metric scores each retrieved unit on a 0--5 scale, evaluating whether the unit is well-supported by the raw retrieved content and relevant to the user's context. The same scoring criteria are applied to both LLM-judged and human evaluation, with the prompt-defined rubric provided below.

\begin{tcolorbox}[
  colback=green!5,
  colframe=green!40,
  breakable,
  title={\textbf{Precision Judge Prompt -- System (English Translation)}},
  fonttitle=\small,
  boxrule=0.5pt,
  arc=2pt,
  left=6pt, right=6pt, top=4pt, bottom=4pt
]
{\small\ttfamily
You are a memory recall evaluation expert. Your task is: given a user message and the raw content actually retrieved by the agent from the memory store, score each ``retrieved unit'' output by the agent on a 0--5 quality scale.\\[6pt]
\textbf{Scoring criteria} (integer, 0--5):\\
0: Fabrication -- the memory unit has no basis in the retrieved content and is entirely made up; or the memory unit can be directly extracted from the user message;\\
1: Reasonable unit name -- the memory unit sounds reasonable but has no clear support in the retrieved content;\\
2: Reasonable path -- association evidence can be found in the retrieved content, and the association path is basically logical;\\
3: Matches trigger condition -- the memory unit is supported by raw content, and the association path explicitly matches the trigger type (temporal/entity/emotional/behavioral pattern/multi-hop);\\
4: High quality -- the memory unit highly aligns with the retrieved content, the path is clear and natural, and strongly relevant to the user's context;\\
5: Excellent quality -- the memory unit is precise and compelling, the path is impeccable, and represents the most valuable memory association for the user's context.\\[6pt]
Output must be in JSON format only, with no extra text.
}
\end{tcolorbox}

\begin{tcolorbox}[
  colback=green!5,
  colframe=green!40,
  breakable,
  title={\textbf{Precision Judge Prompt -- User (English Translation)}},
  fonttitle=\small,
  boxrule=0.5pt,
  arc=2pt,
  left=6pt, right=6pt, top=4pt, bottom=4pt
]
{\small\ttfamily
\textbf{Trigger type}: \{trigger\_type\} (\{desc\})\\[4pt]
\textbf{User message}: \{question\}\\[6pt]
\textbf{Raw content actually retrieved by the agent from the memory store}:\\
\{context\_str\}\\[6pt]
\textbf{List of retrieved memory units output by the agent}:\\
\{retrieved\_str\}\\[6pt]
\textbf{Please score each memory unit on the 0--5 scale and provide justification}.\\[4pt]
\textbf{Please output in the following JSON format}:\\
\{\\
\hspace*{1em}``judgments'': {[}\\
\hspace*{2em}\{\\
\hspace*{3em}``memory\_unit'': ``memory unit name'',\\
\hspace*{3em}``score'': 0 to 5,\\
\hspace*{3em}``reason'': ``scoring justification''\\
\hspace*{2em}\}\\
\hspace*{1em}{]},\\
\hspace*{1em}``overall\_comment'': ``overall evaluation of retrieval quality (1--2 sentences)''\\
\}
}
\end{tcolorbox}

\noindent\textbf{Variable Descriptions:}
\begin{itemize}
    \item \texttt{\{trigger\_type\}}: The type of memory trigger (e.g., temporal, entity-related, emotional, behavioral pattern, or multi-hop).
    \item \texttt{\{desc\}}: A brief description explaining what the trigger type refers to in terms of memory association.
    \item \texttt{\{question\}}: The user's input message.
    \item \texttt{\{context\_str\}}: The concatenated raw content from all tool calls made by the agent during retrieval, serving as the evidence base for judgment.
    \item \texttt{\{retrieved\_str\}}: The JSON-formatted list of retrieved memory units output by the agent, each containing a memory unit name and an association reason.
\end{itemize}

\subsection{Human Validation}
\label{app:human-validation}

To verify the quality of our automatically constructed benchmark, we conduct a human spot-check on a random sample of 125 test instances (25 per topic domain). Two annotators independently evaluate each instance along three criteria: (1)~\textbf{query validity}---whether the trigger utterance is natural and contains only the intended trigger memory unit without leaking candidate information; (2)~\textbf{candidate correctness}---whether each candidate in the ground-truth set is genuinely inferable from the conversation history and absent from the query; and (3)~\textbf{evidence support}---whether the cited source memory units fully support the association path described in the reasoning field.

An instance is marked as \emph{accepted} if it passes all three criteria, and \emph{rejected} otherwise. Table~\ref{tab:human-validation} summarizes the results.

\begin{table}[h]
\centering
\small
\begin{tabular}{lc}
\toprule
\textbf{Criterion} & \textbf{Pass Rate} \\
\midrule
Query validity & 100\% \\
Candidate correctness & 98.4\% \\
Evidence support & 98.4\% \\
\midrule
Overall acceptance & 98.4\% \\
\bottomrule
\end{tabular}
\caption{Human validation results on randomly sampled instances.}
\label{tab:human-validation}
\end{table}

The human spot-check shows that accepted instances have reliable queries and source-aligned quotations. The remaining invalid cases are primarily \emph{evidence-support errors}, where the quoted evidence is related to the reference answer but does not fully support all required claims. We attribute this to the multi-hop nature of hard-difficulty instances, where the LLM-generated reasoning occasionally omits an intermediate link. Based on these findings, we manually corrected the rejected instances and incorporated the revised versions into the final benchmark.

\subsection{LLM-based candidate completeness check.}
\begin{figure*}[t]
    \centering
    \includegraphics[width=\linewidth]{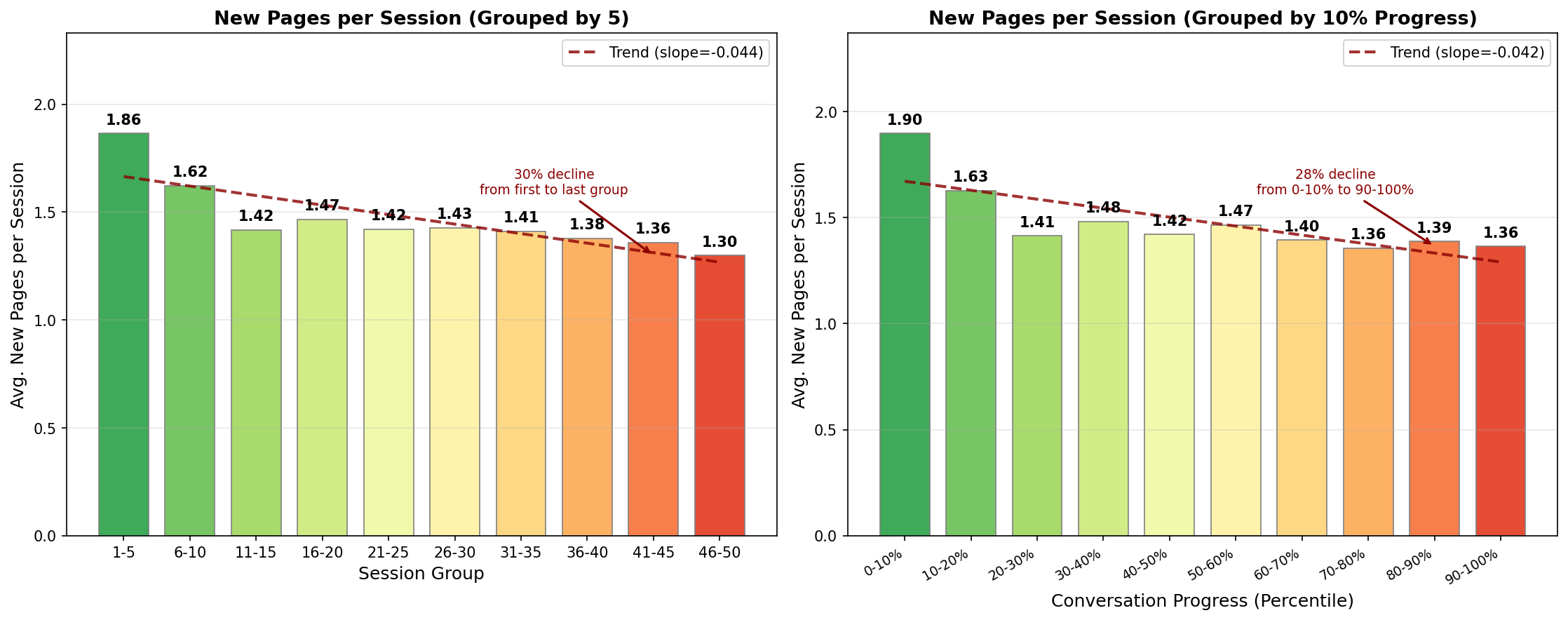}
    \caption{New pages created per session as conversations progress. Left: grouped by every 5 sessions; Right: grouped by 10\% conversation progress percentile. Both views exhibit a consistent declining trend with negative slope, indicating sub-linear page growth.}
    \label{fig:scaling_trend}
\end{figure*}

Since each test instance only retains the top-3 most relevant memory units as its ground-truth candidate set, a natural concern is whether important candidates are omitted. To verify completeness, we employ three strong LLMs (GPT-4o, Claude-3.5-Sonnet, and Gemini-1.5-Pro) as independent judges. For each test instance, we provide the full conversation history and the trigger utterance, and ask each model to list \emph{all} memory units it deems worth proactively surfacing, ranked by relevance. We then compare each model's top-3 selections against our ground-truth candidate set. A candidate is flagged as a potential omission if at least two out of three judges rank it within their top-3 but it is absent from our annotations.

Results show that across all 200 instances, the three judges reach majority agreement with our ground-truth candidate set on 96\% of instances (i.e., at least 2 of 3 candidates overlap). For the remaining 4\%, manual inspection reveals that the divergence stems from equally plausible alternative associations rather than clear omissions---confirming that our top-3 selection is well-calibrated and the ground-truth candidate sets are reliable.

\subsection{Human Evaluation Details}
\label{appendix:human-eval}

\begin{table*}[ht]
\centering
\scriptsize
\caption{Navigation strategies across query complexity levels. Each category illustrates how the agent adapts its navigation depth and tool selection to the task's demands.}
\label{tab:navigation_strategies}
\renewcommand{\arraystretch}{1.4}
\begin{tabularx}{\textwidth}{@{}p{1.8cm} p{3.2cm} p{1.8cm} X@{}}
\toprule
\textbf{Category} & \textbf{Example Query} & \textbf{Case Label} & \textbf{Navigation Route \& Strategy} \\
\midrule
Simple Factual & What degree did I graduate with? & single-session-user & [1--2 steps] \texttt{browse} $\to$ \texttt{list} \newline Targeted lookup; agent recognizes low complexity and selects minimal depth. \\
\addlinespace
Enumeration & How many pieces of furniture did I buy, assemble, sell, or fix? & multi-session & [3--5 steps] \texttt{browse} $\to$ \texttt{list} $\to$ \texttt{read\_page}$\times N$ $\to$ \texttt{follow\_links} \newline Exhaustive traversal; iterates until no further relevant content remains. \\
\addlinespace
Comparative & Which social media platform did I gain the most followers on? & multi-session & [3--4 steps] \texttt{browse} $\to$ \texttt{list} $\to$ \texttt{read\_page}$\times N$ \newline Reads multiple pages to extract comparable metrics for structured comparison. \\
\addlinespace
Timeline Reconstruction & Which task did I complete first, fixing the fence or trimming the goats' hooves? & temporal-reasoning & [3--4 steps] \texttt{browse} (multi-dim) $\to$ \texttt{list} $\to$ \texttt{read\_page}$\times N$ \newline Browses multiple dimensions; extracts and compares temporal markers. \\
\addlinespace
Open-ended Exploration & Can you suggest some activities I can do during my commute? & single-session-preference & [2--4 steps] \texttt{browse} $\to$ \texttt{list} $\to$ \texttt{read\_page}$\times N$ \newline Hierarchical browsing; surveys broadly then synthesizes personalized recommendations. \\
\bottomrule
\end{tabularx}
\end{table*}

We randomly recruited student volunteers from our institution to evaluate model outputs. Annotators were provided with the same scoring rubric used for our LLM-based evaluation and asked to rate outputs from all baselines on a Likert scale. Compensation was mutually negotiated and agreed upon prior to the task. All participants were informed that their ratings would be used for model output evaluation in this research, and they voluntarily agreed to participate. Our human evaluation involves minimal risk: annotators only rate model outputs on a Likert scale, no sensitive or personal information is collected, and all participants are voluntary adults who were informed of the task purpose and compensated fairly. Our institution does not require formal approval for such low-risk annotation tasks, and the study was determined exempt from full ethics review under our institutional guidelines. Annotators were randomly selected and their demographic characteristics have negligible influence on the scoring results, as the task solely involves rating model output quality on predefined criteria.

\section{Page Growth Scaling Analysis}
\label{sec:appendix_trend}

To empirically validate the Page growth management described in \S\ref{sec:scalability}, we track the number of newly created pages per session as conversations progress across multiple long-conversation indices. We analyze this scaling behavior under two complementary groupings: fixed-window (every 5 sessions) and proportional-progress (every 10\% of total sessions).

As shown in Figure~\ref{fig:scaling_trend}, the average number of new pages per session declines monotonically over time. In the fixed-window view (left), early sessions (1--5) average 1.86 new pages per session, which drops to 1.30 by sessions 46--50, representing a 30\% overall decline (trend slope $= -0.044$). The proportional-progress view (right) corroborates this pattern: the rate decreases from 1.90 pages/session in the first 10\% to 1.36 in the final 10\%, a 28\% decline (slope $= -0.042$). Notably, the decline is steepest in the initial phase (approximately the first 15 sessions or 20\% of progress), after which the rate stabilizes around 1.3--1.4 pages/session.

This sub-linear scaling behavior reflects a natural convergence property of the \navimemstruc{}: as the knowledge base accumulates, incoming information increasingly maps onto existing pages through fact updates and section merges rather than triggering new page creation. The early rapid growth corresponds to establishing the user's core knowledge dimensions, while the subsequent plateau indicates that the memory structure has reached sufficient coverage and primarily absorbs new information into its existing organization. This empirical evidence confirms that the \navimemstruc{} does not grow unboundedly, alleviating concerns about long-term page bloat.

\section{Case Study: Navigation as Reasoning}
\label{appendix_navigation}

The Cross-Dimensional  Navigation Interface transforms memory retrieval into an integral part of the reasoning process. Within a standard ReAct loop, the agent interleaves thinking and navigation: each Think step evaluates whether current information is sufficient and plans the next navigation action; each Act step invokes a navigation tool; each Obs step provides structured results that feed back into reasoning. This tight coupling fundamentally distinguishes multi-step navigation from single-shot retrieval across a spectrum of query complexities.

Table~\ref{tab:navigation_strategies} presents cases selected from LongMemEval~\cite{longmemeval}. The corresponding navigation trace examples across five categories are detailed below. For clarity, we only list selected excerpts from the traces that highlight the most critical clues.

\paragraph{Simple Factual Queries.}
For simple factual queries that target a coherent knowledge unit (e.g., ``What degree did I graduate with?''), a targeted \texttt{browse\_dimension} or \texttt{read\_page} call retrieves comprehensive information in one to two steps. The agent autonomously recognizes the low complexity and selects minimal navigation depth: browsing the topic dimension to locate the answer directly in the high-level summary, then optionally confirming via a page listing---without deeper exploration. However, as query complexity increases, the limitations of such single-step retrieval become apparent.

\begin{tcolorbox}[colback=yellow!5, colframe=yellow!50, coltitle=black, breakable, title=Simple Factual Query Example: Single-Step Navigation]
\small
\textbf{Question:} \textit{``What degree did I graduate with?''}\\
\textbf{Ground-truth Answer:} Business Administration

\medskip
\textbf{Step 1: Browse Dimension} $\to$ \texttt{topic}\\
The agent browses the topic dimension and immediately finds the answer in the high-level summary: ``a Business Administration degree.''

\begin{lstlisting}[style=grayoutput]
### topic
- The user is a senior UX and motion designer, lecturer, and craft vendor with a Business Administration degree who enjoys strategic logic puzzles, D&D, sci-fi movies, live music, and various sports, pursues fitness through a Fitbit, morning jogs, yoga, foam rolling, cycling, and tennis while seeking natural remedies, mindfulness practices, and consistent routines, relieves stress with a lavender-chamomile diffuser, monitors blood pressure, intends to get a flu shot, professionally requests lecture plans on radiation delivery systems and conducts augmented reality marketing research, creatively rewrites movie scripts into crossovers avoiding cliches, explores Python to plot script intensity and build interactive stock charts with alarms and stop-losses, analyzes DISC personality conflicts, considers
\end{lstlisting}

\medskip
\textbf{Step 2: List Dimension} $\to$ \texttt{user/assistant/topic/index.md}\\
The agent confirms via the page listing, which shows an \texttt{education} entry explicitly stating ``graduated with a degree in Business Administration.''

\begin{lstlisting}[style=grayoutput]
[user/assistant/topic/index.md]
# topic
- The user is a senior UX and motion designer, lecturer, and craft vendor with a Business Administration degree who enjoys strategic logic puzzles, D&D, sci-fi movies, live music, and various sports, pursues fitness through a Fitbit, morning jogs, yoga, foam rolling, cycling, and tennis while seeking natural remedies, mindfulness practices, and consistent routines, relieves stress with a lavender-chamomile diffuser, monitors blood pressure, intends to get a flu shot, professionally requests lecture plans on radiation delivery systems and conducts augmented reality marketing research, creatively rewrites movie scripts into crossovers avoiding cliches, explores Python to plot script intensity and build interactive stock charts with alarms and stop-losses, analyzes DISC personality conflicts, considers

## Pages
- [[education]] (degree, college) : User graduated with a degree in Business Administration, which has helped them in their new role. #Business Administration #graduation
- ...
\end{lstlisting}

\medskip
\textbf{Prediction:}\\
Business Administration \checkmark
\end{tcolorbox}

\paragraph{Enumeration Tasks.}
Enumeration tasks such as ``How many pieces of furniture did I buy, assemble, sell, or fix?'' require exhaustive coverage that a single query cannot guarantee. Multi-step navigation through three to five actions is warranted: the agent browses the relevant dimension, lists sub-pages to identify candidates, reads multiple pages to collect all matching evidence, and follows cross-dimensional associative links to verify completeness---iterating over entries systematically until no further relevant content remains.

\begin{tcolorbox}[colback=yellow!5, colframe=yellow!50, coltitle=black, breakable, title=Enumeration Task Example: Multi-Step Navigation]
\small
\textbf{Question:} \textit{``How many pieces of furniture did I buy, assemble, sell, or fix in the past few months?''}\\
\textbf{Ground-truth Answer:} 4

\medskip
\textbf{Step 1: Browse Dimension} $\to$ \texttt{topic}\\
The agent first browses the topic dimension and identifies a high-level summary mentioning home-related activities.

\begin{lstlisting}[style=grayoutput]
### topic
- The user is a software sales professional utilizing MEDDPICC and Obsidian who enjoys autocross with their modified 2018 Honda Civic Si, farming chickens, pigs, and a goat, caring for their kitten Luna and older dog Max, pursuing diverse artistic hobbies like Saturday sculpture classes, model building, photography with vintage cameras, playing guitar and violin, visiting art exhibitions, and listening to art and music podcasts, while conducting a systematic review on biochemical sensors for neuromodulation, writing a thesis on social media marketing, exploring Stoicism, feminism, protest and indie music, e-waste regulations, the Dictionary of Canadian Biography, and democratic election processes, and actively managing home maintenance, decor upgrades, healthier cooking and baking, travel planning, cryptocurrency security,
\end{lstlisting}

\medskip
\textbf{Step 2: List Dimension} $\to$ \texttt{user/assistant/topic/index.md}\\
The agent lists all sub-pages under the topic dimension, identifying relevant entries: \texttt{home decor}, \texttt{home improvement}, \texttt{home office}, \texttt{dog care}, \texttt{cat care}.

\begin{lstlisting}[style=grayoutput]
[user/assistant/topic/index.md]
# topic
- The user is a software sales professional utilizing MEDDPICC and Obsidian who enjoys autocross with their modified 2018 Honda Civic Si, farming chickens, pigs, and a goat, caring for their kitten Luna and older dog Max, pursuing diverse artistic hobbies like Saturday sculpture classes, model building, photography with vintage cameras, playing guitar and violin, visiting art exhibitions, and listening to art and music podcasts, while conducting a systematic review on biochemical sensors for neuromodulation, writing a thesis on social media marketing, exploring Stoicism, feminism, protest and indie music, e-waste regulations, the Dictionary of Canadian Biography, and democratic election processes, and actively managing home maintenance, decor upgrades, healthier cooking and baking, travel planning, cryptocurrency security,

## Pages
- [[bathroom maintenance]] : User is experiencing a slow-draining bathroom sink and has used a plunger multiple times. They plan to buy a mesh screen to prevent future clogs. They also cleaned their shower curtain on the weekend of May 13-14, 2023, scrubbing off soap scum and mildew. Additionally, they plan to clean the sink ba #home maintenance #cleaning #plumbing
- [[home decor]] : The user is upgrading their home decor by rearranging their living room, purchasing a new metal-legged coffee table, and shopping for light gray or beige throw pillows with wooden or metal accents to match a future sectional sofa, alongside a modern table lamp with metallic accents #living room #bedroom #furniture
- [[cat care]] : User is actively cat-proofing their home for their new kitten, Luna. On 2023-05-23, they bought scratch guards from IKEA to protect furniture from Luna's scratching, which has been effective. They also plan to rotate Luna's cat tree around the house every few days for a change of scenery. User has e #pet care #cat-proofing #grooming
- [[home improvement]] : User fixed a wobbly leg on their kitchen table the weekend before 2023-05-26 by tightening a screw with a screwdriver; the wobbly leg had been bothering them for months. User is also planning to reorganize kitchen cabinets to improve flow and maximize storage space for cooking utensils and gadgets. #DIY #kitchen #furniture repair
- [[dog care]] : As of 2023-05-28, user is planning to get a new orthopedic dog bed for Max, who is getting older and needs joint comfort. User is leaning towards the Big Barker brand due to its high quality, 10-year warranty, and 100-night sleep trial. User also plans to wash Max's old blankets and beds to remove l #dog bed #orthopedic #Big Barker
- [[home office]] : User assembled an IKEA bookshelf for their home office around March 2023 (about two months prior to May 29, 2023). The bookshelf has been a game-changer for their productivity, helping them stay organized and focused. #IKEA #bookshelf #organization
- ...
\end{lstlisting}

\medskip
\textbf{Step 3: Read Pages} $\to$ \texttt{home decor.md}, \texttt{home improvement.md}, \texttt{home office.md}, \texttt{dog care.md}, \texttt{cat care.md}\\
The agent reads multiple pages to collect furniture-related events:
\begin{itemize}[nosep, leftmargin=1.5em]
    \item \texttt{home decor}: Bought a West Elm wooden coffee table with metal legs (delivered May 18, 2023); ordered a Casper mattress (May 14, 2023).
    \item \texttt{home improvement}: Fixed a wobbly kitchen table leg (weekend before May 26, 2023).
    \item \texttt{home office}: Assembled an IKEA bookshelf ($\sim$March 2023).
\end{itemize}

\begin{lstlisting}[style=grayoutput]
[user/assistant/topic/home decor.md]
# home decor
- The user is upgrading their home decor by rearranging their living room, purchasing a new metal-legged coffee table, and shopping for light gray or beige throw pillows with wooden or metal accents to match a future sectional sofa, alongside a modern table lamp with metallic accents
- aliases: []
- tags: [living room, bedroom, furniture, shopping]

## Home decoration and furniture upgrades
- category: experience
- detail: User is upgrading their home decor and furniture. They rearranged their living room and got a new coffee table with metal legs. They are looking for light gray or beige throw pillows with wooden or metal accents to match a future sectional sofa. They also want a modern table lamp with metallic accents to match the coffee table's metal legs. For the bedroom, they ordered a new Casper mattress on 2023-05-14 (referenced as last week), which is scheduled to arrive on 2023-05-24 (referenced as next Wednesday). They are also looking for bedside tables with metal or glass accents to complement the Casper mattress and their modern aesthetic.
- Page:
  - [[user/assistant/anniversary/casper mattress delivery.md]]

## Shopping for throw pillows and home decor preferences
- category: preference
- detail: User is shopping for 20" x 20" throw pillows for a standard 3-seater couch, starting with two pillows. They prefer shopping at West Elm, having previously bought a wooden coffee table with metal legs from them. When choosing throw pillows, they consider the couch fabric, wall color, and the industrial aesthetic of their coffee table's metal legs. They are looking for patterns and designs that tie in with the wood and metal aesthetic.
- Page:
  - [[user/assistant/topic/home office.md]]

[user/assistant/topic/home improvement.md]
# home improvement
- User fixed a wobbly leg on their kitchen table the weekend before 2023-05-26 by tightening a screw with a screwdriver; the wobbly leg had been bothering them for months. User is also planning to reorganize kitchen cabinets to improve flow and maximize storage space for cooking utensils and gadgets.
- aliases: []
- tags: [DIY, kitchen, furniture repair, organization]

## Home improvement and organization activities
- category: experience
- detail: User fixed a wobbly leg on their kitchen table the weekend before 2023-05-26 by tightening a screw with a screwdriver; the wobbly leg had been bothering them for months. User is also planning to reorganize kitchen cabinets to improve flow and maximize storage space for cooking utensils and gadgets.
- Page:
  - [[user/assistant/topic/interior design.md]]

[user/assistant/topic/home office.md]
# home office
- User assembled an IKEA bookshelf for their home office around March 2023 (about two months prior to May 29, 2023). The bookshelf has been a game-changer for their productivity, helping them stay organized and focused.
- aliases: []
- tags: [IKEA, bookshelf, organization, productivity]

## Assembled IKEA bookshelf for home office
- category: experience
- detail: User assembled an IKEA bookshelf for their home office around March 2023 (about two months prior to May 29, 2023). The bookshelf has been a game-changer for their productivity, helping them stay organized and focused.
- Page:
  - [[user/assistant/topic/home decor.md]]

[user/assistant/topic/dog care.md]
# dog care
- As of 2023-05-28, user is planning to get a new orthopedic dog bed for Max, who is getting older and needs joint comfort. User is leaning towards the Big Barker brand due to its high quality, 10-year warranty, and 100-night sleep trial. User also plans to wash Max's old blankets and beds to remove l
- aliases: []
- tags: [dog bed, orthopedic, Big Barker, pet care]

## Looking to buy an orthopedic dog bed for Max
- category: experience
- detail: As of 2023-05-28, user is planning to get a new orthopedic dog bed for Max, who is getting older and needs joint comfort. User is leaning towards the Big Barker brand due to its high quality, 10-year warranty, and 100-night sleep trial. User also plans to wash Max's old blankets and beds to remove lingering scents before introducing the new bed.
- Page:
  - [[user/assistant/topic/cooking.md]]
  - [[user/assistant/figure/max.md]]

[user/assistant/topic/cat care.md]
# cat care
- User is actively cat-proofing their home for their new kitten, Luna. On 2023-05-23, they bought scratch guards from IKEA to protect furniture from Luna's scratching, which has been effective. They also plan to rotate Luna's cat tree around the house every few days for a change of scenery. User has e
- aliases: []
- tags: [pet care, cat-proofing, grooming]

## Cat-proofing and grooming kitten Luna
- category: experience
- detail: User is actively cat-proofing their home for their new kitten, Luna. On 2023-05-23, they bought scratch guards from IKEA to protect furniture from Luna's scratching, which has been effective. They also plan to rotate Luna's cat tree around the house every few days for a change of scenery. User has established a grooming routine for Luna, brushing her 2-3 times a week for 5-10 minutes using a soft-bristle kitten brush. Luna initially squirmed but now enjoys and looks forward to brushing sessions, resulting in a healthier, shinier coat and increased affection. User is considering introducing nail trimming and ear cleaning to the routine soon.
- Page:
  - [[user/assistant/figure/luna.md]]
\end{lstlisting}

\medskip
\textbf{Step 4: Follow Links} $\to$ \texttt{interior design.md}\\
The agent follows cross-dimensional associative links to verify no additional furniture events are missed.

\begin{lstlisting}[style=grayoutput]
[user/assistant/topic/interior design.md]
# interior design
- A user updating their modern living room purchased a West Elm wooden coffee table with metal legs and wants to incorporate navy accents, specifically liking the Navy Grid Velvet Pillow Cover and other throw pillows
- aliases: []
- tags: [home decor, living room, modern aesthetic, navy accents]

## Modern living room decor and navy accent preferences
- category: preference
- detail: User enjoys modern interior design and is updating their living room. They purchased a wooden coffee table with metal legs from West Elm about three weeks before May 26, 2023 (delivered on May 18, 2023). They like the Navy Grid Velvet Pillow Cover from West Elm and want to add navy accents to their living room for a pop of color. They prefer to test colors with smaller accents first, like a navy throw blanket, before committing to larger items like a navy area rug.

## Living room decoration and throw pillow preferences
- category: preference
- detail: User has a modern living room with a neutral color scheme (beige, gray, white) and a wooden coffee table with metal legs. On 2023-05-26, user was looking for throw pillows to complement the modern feel and add a pop of color. User likes velvet pillows in rich hues, specifically mustard and teal. Prefers to start with one bold, rich colored pillow and order swatches/samples before making a final decision. Interested in options from West Elm and CB2.
- Page:
  - [[user/assistant/topic/home improvement.md]]
\end{lstlisting}

\medskip
\textbf{Prediction:}\\
\texttt{1. \textbf{Bought} -- West Elm wooden coffee table with metal legs (delivered May 18, 2023)}\\
\texttt{2. \textbf{Bought} -- Casper mattress (ordered May 14, 2023; delivered May 24, 2023)}\\
\texttt{3. \textbf{Assembled} -- IKEA bookshelf for home office ($\sim$March 2023)}\\
\texttt{4. \textbf{Fixed} -- Kitchen table wobbly leg (weekend before May 26, 2023)}\\
No records of selling any furniture. \textbf{Answer: 4} \checkmark
\end{tcolorbox}

\paragraph{Comparative Tasks.}
Comparative tasks such as ``Which social media platform did I gain the most followers on?'' demand synthesizing quantitative information from multiple distinct pages. While single-shot retrieval might return partial results requiring post-hoc assembly, multi-step navigation through three to four actions enables the agent to browse the relevant dimension, identify candidate pages via listing, read each page to extract comparable metrics, and perform structured comparison with full context.

\begin{tcolorbox}[colback=yellow!5, colframe=yellow!50, coltitle=black, breakable, title=Comparative Task Example: Multi-Step Navigation]
\small
\textbf{Question:} \textit{``Which social media platform did I gain the most followers on over the past month?''}\\
\textbf{Ground-truth Answer:} TikTok

\medskip
\textbf{Step 1: Browse Dimension} $\to$ \texttt{topic}\\
The agent browses the topic dimension and identifies social media--related activities in the user's profile.

\begin{lstlisting}[style=grayoutput]
### topic
- The user is a vegan Year 10 student and non-profit leader who combats burnout through self-care, balancing a structured morning routine of journaling, freewriting, and a 2,000-word daily writing goal with poetry, short stories, and promotional writing for Gary Dranow and the Manic Emotions, while advocating for mental health and music therapy, organizing culturally sensitive social justice programs and innovative donor campaigns using magic events and VR/AR, exploring permaculture and sustainable living, investing in real estate and mountain cabins, enjoying skincare, home mixology, vegan meal prep, crochet, crafts, photography, yoga, mythology, Harry Potter, and Mandarin, managing a daily routine and bedtime reading, parenting bilingual children with speech concerns
\end{lstlisting}

\medskip
\textbf{Step 2: List Dimension} $\to$ \texttt{user/assistant/topic/index.md}\\
The agent lists sub-pages and identifies three relevant entries: \texttt{social media}, \texttt{social media marketing}, and \texttt{mental health awareness campaign}.

\begin{lstlisting}[style=grayoutput]
[user/assistant/topic/index.md]
# topic
- The user is a vegan Year 10 student and non-profit leader who combats burnout through self-care, balancing a structured morning routine of journaling, freewriting, and a 2,000-word daily writing goal with poetry, short stories, and promotional writing for Gary Dranow and the Manic Emotions, while advocating for mental health and music therapy, organizing culturally sensitive social justice programs and innovative donor campaigns using magic events and VR/AR, exploring permaculture and sustainable living, investing in real estate and mountain cabins, enjoying skincare, home mixology, vegan meal prep, crochet, crafts, photography, yoga, mythology, Harry Potter, and Mandarin, managing a daily routine and bedtime reading, parenting bilingual children with speech concerns

## Pages
- [[social media marketing]] : User attended a workshop on social media marketing on 2023-05-21. They also mentioned attending a similar workshop about two weeks prior to 2023-05-21 (around early May 2023), where they met interesting people, including a woman working for a non-profit that uses social media to drive social change. #social media #marketing #workshop
- [[mental health awareness campaign]] (#BreakTheSilenceMH) : User is creating a social media campaign to raise awareness about mental health, an issue close to their heart. They chose the branded hashtag #BreakTheSilenceMH for the campaign. They plan to share a series of posts on Instagram and Twitter, and are creating a content calendar to track posts. They #mental health #social media campaign #awareness
- [[social media]] : An active social media content creator across multiple platforms, the user focuses on optimizing their online presence and driving website traffic through a Twitter growth strategy of consistently posting 5-7 daily educational and personal tweets without a set content calendar #Twitter #social media strategy #online presence
- ...
\end{lstlisting}

\medskip
\textbf{Step 3: Read Pages} $\to$ \texttt{social media.md}, \texttt{social media marketing.md}, \texttt{mental health awareness campaign.md}\\
The agent reads the pages and compares follower growth across platforms:
\begin{itemize}[nosep, leftmargin=1.5em]
    \item \texttt{social media}: TikTok gained $\sim$200 followers in three weeks; Twitter grew from 420 to 540 (+120) over the past month; Facebook steady at $\sim$800.
    \item \texttt{social media marketing}: Attended workshops; no follower data.
    \item \texttt{mental health awareness campaign}: Campaign on Instagram and Twitter; no follower counts.
\end{itemize}

\begin{lstlisting}[style=grayoutput]
[user/assistant/topic/social media.md]
# social media
- An active social media content creator across multiple platforms, the user focuses on optimizing their online presence and driving website traffic through a Twitter growth strategy of consistently posting 5-7 daily educational and personal tweets without a set content calendar
- aliases: []
- tags: [Twitter, social media strategy, online presence, engagement]

## User's Twitter Strategy and Growth
- category: experience
- detail: User is focused on optimizing their social media strategy to increase online presence and drive traffic to their website. On Twitter, they post a mix of educational and personal content, aiming for 5-7 tweets per day. They do not have a specific content calendar but try to stay consistent and engage with followers by responding to comments and DMs. Over the past month (as of 2023-05-29), their Twitter follower count grew from 420 to 540. User wants to increase their engagement rate and is interested in hosting a Twitter Chat or Q&A session, having participated in Twitter Chats before.

## Active social media content creator across multiple platforms
- category: experience
- detail: User is active on Instagram, Facebook, Twitter, and TikTok, posting regularly about hobbies, interests, daily life, and updates about favorite TV shows and movies. Their goals are to drive engagement and increase follower count across all platforms. By 2023-05-29, they had gained around 200 followers on TikTok over the past three weeks. They get a lot of views on their Instagram stories and plan to use Instagram's 'Question' sticker to ask followers about their favorite TV shows and movies to encourage engagement.
- Page:
  - [[user/assistant/topic/tv and movies.md]]

## Active social media user across multiple platforms
- category: experience
- detail: As of May 30, 2023, user is actively working on improving social media engagement across Instagram, Facebook, and TikTok. On Instagram, they are doing pretty well and their stories get a lot of views, especially when using the 'question' sticker. On Facebook, their follower count has remained steady at around 800, but their posts have been getting more shares and comments than usual. On TikTok, they post short videos showcasing their hobbies and interests, and have been surprised by how quickly their follower count has grown. User is interested in learning the best times to post, how to create engaging content, and strategies to grow their following on these platforms.

## User's Twitter usage and social media strategy
- category: experience
- detail: User is focused on optimizing their social media strategy to increase online presence and drive traffic to their website. On Twitter, they post a mix of educational and personal content, aiming for 5-7 tweets per day. They do not use a specific content calendar but try to stay consistent and engage with followers by responding to comments and DMs. Over the past month (as of 2023-05-29), their Twitter follower count grew from 420 to 540. They have participated in Twitter Chats before and are interested in hosting their own Twitter Chat or Q&A session to increase engagement, and are currently seeking tips on how to promote it.

## Active social media content creator across multiple platforms
- category: experience
- detail: User is active on Instagram, Facebook, Twitter, and TikTok, posting regularly about hobbies, interests, daily life, and updates about favorite TV shows and movies. Goals include driving engagement and increasing follower count across all platforms. By 2023-05-29, user had gained around 200 followers on TikTok over the past three weeks. User gets a lot of views on Instagram stories and plans to use Instagram's 'Question' sticker to ask followers about their favorite TV shows and movies to encourage engagement.
- Page:
  - [[user/assistant/topic/reading.md]]

## Social media usage and engagement strategy
- category: experience
- detail: As of 2023-05-30, user is actively working on improving social media engagement across multiple platforms. On Instagram, they are doing pretty well and their stories get a lot of views, especially when using the 'question' sticker. On Facebook, their follower count has remained steady at around 800, but their posts have been getting more shares and comments than usual. On TikTok, they post short videos showcasing their hobbies and interests, and have been surprised by how quickly their follower count has grown. They are looking for advice on the best times to post, how to create engaging content, and how to grow their following, particularly on Facebook and TikTok.

[user/assistant/topic/social media marketing.md]
# social media marketing
- User attended a workshop on social media marketing on 2023-05-21. They also mentioned attending a similar workshop about two weeks prior to 2023-05-21 (around early May 2023), where they met interesting people, including a woman working for a non-profit that uses social media to drive social change.
- aliases: []
- tags: [social media, marketing, workshop]

## Attended a social media marketing workshop
- category: experience
- detail: User attended a workshop on social media marketing on 2023-05-21. They also mentioned attending a similar workshop about two weeks prior to 2023-05-21 (around early May 2023), where they met interesting people, including a woman working for a non-profit that uses social media to drive social change. They exchanged numbers and user has been following the organization's social media accounts since then.
- Page:
  - [[user/assistant/topic/mental health awareness campaign.md]]
  - [[user/assistant/figure/non-profit woman from workshop.md]]

[user/assistant/topic/mental health awareness campaign.md]
# mental health awareness campaign
- User is creating a social media campaign to raise awareness about mental health, an issue close to their heart. They chose the branded hashtag #BreakTheSilenceMH for the campaign. They plan to share a series of posts on Instagram and Twitter, and are creating a content calendar to track posts. They
- aliases: [#BreakTheSilenceMH]
- tags: [mental health, social media campaign, awareness]

## Creating a mental health awareness social media campaign
- category: experience
- detail: User is creating a social media campaign to raise awareness about mental health, an issue close to their heart. They chose the branded hashtag #BreakTheSilenceMH for the campaign. They plan to share a series of posts on Instagram and Twitter, and are creating a content calendar to track posts. They previously shared personal stories about mental health on Instagram a few weeks prior to 2023-05-21.
- Page:
  - [[user/assistant/topic/social media marketing.md]]
  - [[user/assistant/figure/non-profit woman from workshop.md]]
\end{lstlisting}

\medskip
\textbf{Prediction:}\\
TikTok --- gained $\sim$200 followers in the past three weeks (as of 2023-05-29), compared to +120 on Twitter over the past month and steady/no growth on Facebook. \checkmark
\end{tcolorbox}

\paragraph{Timeline Reconstruction Tasks.}
Timeline reconstruction tasks such as ``Which task did I complete first, fixing the fence or trimming the goats' hooves?'' suffer from the inherent orderlessness of retrieval results. Multi-step navigation through three to four actions resolves this: the agent browses multiple dimensions (e.g., topic and place) to locate relevant entities, reads specific pages to extract temporal markers, and compares dates to establish chronological order---preserving coherence that single-shot retrieval cannot guarantee.

\begin{tcolorbox}[colback=yellow!5, colframe=yellow!50, coltitle=black, breakable, title=Timeline Reconstruction Task Example: Multi-Step Navigation]
\small
\textbf{Question:} \textit{``Which task did I complete first, fixing the fence or trimming the goats' hooves?''}\\
\textbf{Ground-truth Answer:} Fixing the fence

\medskip
\textbf{Step 1: Browse Dimension} $\to$ \texttt{topic}, \texttt{place}\\
The agent browses two dimensions and identifies farm-related content mentioning both goats and fence repair.

\begin{lstlisting}[style=grayoutput]
### topic
- A third-year electrical engineering student and farmer manages diverse interests including Indian agricultural politics, a farm open day with a petting zoo and dairy cow investment, Dublin and Bali historical tours and illuminated manuscripts with their well-behaved pet goat, a healthier baked latke Hanukkah dinner with caramelized onions and braised red cabbage followed by a synagogue candlelight vigil, decade-based road trip playlists balancing upbeat and slow tracks, innovative Pythagorean theorem objective questions in LaTeX, HVAC engineering using Malaysian tariffs, optimizing a 90-minute bus commute with self-improvement podcasts, French economic history in Guise and dining at Le Bocage, casual warm-toned fashion, an FC Barcelona Reddit strategy, Asian fusion cooking and spicy homemade kimchi, applying

### place
- A Toronto Metropolitan University student who runs a goat and chicken farm-planning to add cows, host an open day, open a small shop, and repair an east fence-discussed Indian political strategies and agricultural reforms, Jamaican reggae and ska origins, Hull's urban redevelopment, the Philippines' furniture exports, robotic mapping research at the University of Freiburg, and the Good Friday Agreement regarding Northern Ireland, expressed interest in Dublin independence movement historical tours with student discounts and dining packages, inquired about Bali's Mount Batur and Uluwatu, attended a local Hanukkah candlelight vigil at a nearby synagogue, requested a chiller system operating cost calculation using the Malaysian daylight electrical tariff, planned a trip to Guise, France,
\end{lstlisting}

\medskip
\textbf{Step 2: List Dimension} $\to$ \texttt{user/assistant/topic/index.md}, \texttt{user/assistant/place/index.md}\\
The agent lists sub-pages across both dimensions and identifies two relevant entries: \texttt{pet goat} (under topic) and \texttt{user's farm} (under place).

\begin{lstlisting}[style=grayoutput]
[user/assistant/topic/index.md]
# topic
- A third-year electrical engineering student and farmer manages diverse interests including Indian agricultural politics, a farm open day with a petting zoo and dairy cow investment, Dublin and Bali historical tours and illuminated manuscripts with their well-behaved pet goat, a healthier baked latke Hanukkah dinner with caramelized onions and braised red cabbage followed by a synagogue candlelight vigil, decade-based road trip playlists balancing upbeat and slow tracks, innovative Pythagorean theorem objective questions in LaTeX, HVAC engineering using Malaysian tariffs, optimizing a 90-minute bus commute with self-improvement podcasts, French economic history in Guise and dining at Le Bocage, casual warm-toned fashion, an FC Barcelona Reddit strategy, Asian fusion cooking and spicy homemade kimchi, applying

## Pages
- [[pet goat]] (goat) : The user has a well-behaved pet goat that likes to try new foods, considers it a great travel companion and hopes to bring it on a Dublin dining tour, regularly trims its hooves with improving skill, and has developed an interest in European pet food regulations #pet #goat
- ...

[user/assistant/place/index.md]
# place
- A Toronto Metropolitan University student who runs a goat and chicken farm-planning to add cows, host an open day, open a small shop, and repair an east fence-discussed Indian political strategies and agricultural reforms, Jamaican reggae and ska origins, Hull's urban redevelopment, the Philippines' furniture exports, robotic mapping research at the University of Freiburg, and the Good Friday Agreement regarding Northern Ireland, expressed interest in Dublin independence movement historical tours with student discounts and dining packages, inquired about Bali's Mount Batur and Uluwatu, attended a local Hanukkah candlelight vigil at a nearby synagogue, requested a chiller system operating cost calculation using the Malaysian daylight electrical tariff, planned a trip to Guise, France,

## Pages
- [[user's farm]] : The user owns and operates a farm raising goats and chickens with plans to add cows, intends to build a small shop selling farm products, will host an open day event, has birdwatching spots nearby, and has an east-side fence that needed repair #farm #shop #residence
- ...
\end{lstlisting}

\medskip
\textbf{Step 3: Read Pages} $\to$ \texttt{pet goat.md}, \texttt{user's farm.md}\\
The agent reads both pages and extracts temporal information:
\begin{itemize}[nosep, leftmargin=1.5em]
    \item \texttt{pet goat}: Goat hoof trimming completed successfully on 2023-05-11.
    \item \texttt{user's farm}: East-side fence repaired on 2023-05-04.
\end{itemize}

\begin{lstlisting}[style=grayoutput]
[user/assistant/topic/pet goat.md]
# pet goat
- The user has a well-behaved pet goat that likes to try new foods, considers it a great travel companion and hopes to bring it on a Dublin dining tour, regularly trims its hooves with improving skill, and has developed an interest in European pet food regulations
- aliases: [goat]
- tags: [pet, goat]

## Owns a pet goat
- category: experience
- detail: The user has a pet goat and considers it a great travel companion. The user says this goat is very well-behaved, likes to try new foods and drinks, and hopes to bring it along on a Dublin dining tour.
- Page:
  - [[user/assistant/topic/Irish independence movement history.md]]
  - [[user/assistant/figure/pet goat.md]]
  - [[user/assistant/place/Dublin.md]]

## Regularly trims goat hooves with improving skill
- category: experience
- detail: The user has been working on trimming their farm animals' hooves more regularly and is becoming more proficient with practice. On 2023-05-11, the user successfully completed a hoof trimming without making a mess and felt proud of the accomplishment.
- Page:
  - [[user/assistant/topic/farm management and open day event.md]]
  - [[user/assistant/figure/pet goat.md]]
  - [[user/assistant/figure/farm cows.md]]
  - [[user/assistant/figure/farm chickens.md]]
  - [[user/assistant/place/user's farm.md]]

## Interest in European pet food regulations
- category: interest
- detail: On 2023-05-25, user asked about whether pet foods require pre-market authorization in Europe, indicating an interest in European pet food regulations and market requirements.

[user/assistant/place/user's farm.md]
# user's farm
- The user owns and operates a farm raising goats and chickens with plans to add cows, intends to build a small shop selling farm products, will host an open day event, has birdwatching spots nearby, and has an east-side fence that needed repair
- aliases: []
- tags: [farm, shop, residence]

## User's farm
- category: objective fact
- detail: The user owns and operates a farm raising goats and chickens, with plans to add cows. The farm will host an open day event, and the user also plans to build a small shop on the farm to sell farm products and dairy.
- Page:
  - [[user/assistant/topic/pet goat.md]]
  - [[user/assistant/topic/farm management and open day event.md]]
  - [[user/assistant/figure/pet goat.md]]
  - [[user/assistant/figure/farm cows.md]]
  - [[user/assistant/figure/farm chickens.md]]

## Birdwatching spots near user's home
- category: objective fact
- detail: The user's home is located in the eastern part of the state, with a nature center (featuring mixed forest and grassland habitat) and a park nearby, both of which are frequent birdwatching spots for the user.
- Page:
  - [[user/assistant/topic/birdwatching and equipment.md]]

## East-side fence repair
- category: objective fact
- detail: The user's farm has an east-side fence where the goats like to graze. The fence was repaired on 2023-05-04.
- Page:
  - [[user/assistant/topic/farm management and open day event.md]]
  - [[user/assistant/topic/Asian fusion cooking.md]]
  - [[user/assistant/figure/pet goat.md]]
\end{lstlisting}

\medskip
\textbf{Prediction:}\\
Fixing the fence. The fence was repaired on 2023-05-04, while the goats' hooves were trimmed on 2023-05-11. \checkmark
\end{tcolorbox}

\paragraph{Open-ended Exploratory Queries.}
Open-ended exploratory queries such as ``Can you suggest some activities I can do during my commute?'' are too vague to formulate as effective search queries. The agent handles these through hierarchical browsing across two to four steps: first surveying broadly via dimension browsing to identify relevant topics (commute mode, audio preferences, well-being habits), then reading multiple pages to synthesize personalized recommendations grounded in the user's established preferences, routines, and context.

\begin{tcolorbox}[colback=yellow!5, colframe=yellow!50, coltitle=black, breakable, title=Open-ended Exploratory Query Example: Multi-Step Navigation]
\small
\textbf{Question:} \textit{``Can you suggest some activities I can do during my commute to work?''}\\
\textbf{Ground-truth Answer:} The user would prefer suggestions related to listening to new podcasts or audiobooks, especially genres beyond true crime or self-improvement, such as history. They may not be interested in activities requiring visual attention, as they are cycling.

\medskip
\textbf{Step 1: Browse Dimension} $\to$ \texttt{topic}\\
The agent browses the topic dimension and identifies commute-related interests including cycling, podcasts, music, and anxiety management.

\begin{lstlisting}[style=grayoutput]
### topic
- Managing work-triggered anxiety and deadlines through mindful breathing, structured morning routines with runs, quick healthy meals, and Thursday squash games, the user commutes via a hybrid bicycle while listening to indie rock and podcasts, exploring dystopian AI-generated art and prompt crafting, AI safety, Alan Watts' philosophy, sci-fi and gothic literature, 19th-century Korean and American history, Beyonce's activism, and ocean conservation, planning Outer Banks and international road trips with curated playlists and packed snacks, a sibling camping trip featuring family photo sharing and cooking competitions, developing Brazilian Jiu-Jitsu training programs and a Flow blockchain-based educational game with NFT markets, enjoying films like Everything Everywhere All at Once and Marvel content while listening to film soundtracks to focus
\end{lstlisting}

\medskip
\textbf{Step 2: List Dimension} $\to$ \texttt{user/assistant/topic/index.md}\\
The agent lists sub-pages and identifies five relevant entries spanning commute mode, audio content, and well-being.

\begin{lstlisting}[style=grayoutput]
[user/assistant/topic/index.md]
# topic
- Managing work-triggered anxiety and deadlines through mindful breathing, structured morning routines with runs, quick healthy meals, and Thursday squash games, the user commutes via a hybrid bicycle while listening to indie rock and podcasts, exploring dystopian AI-generated art and prompt crafting, AI safety, Alan Watts' philosophy, sci-fi and gothic literature, 19th-century Korean and American history, Beyonce's activism, and ocean conservation, planning Outer Banks and international road trips with curated playlists and packed snacks, a sibling camping trip featuring family photo sharing and cooking competitions, developing Brazilian Jiu-Jitsu training programs and a Flow blockchain-based educational game with NFT markets, enjoying films like Everything Everywhere All at Once and Marvel content while listening to film soundtracks to focus

## Pages
- [[bicycle commute]] : The user plans to buy a hybrid bicycle for weekday rush-hour commuting to avoid crowded Monday buses. Starting from 123 Oak St to 456 Main St. Prefers bike lanes and quiet streets, avoiding busy roads and steep hills. Planned route is approximately 2.6 miles, estimated 35-45 minutes during morning rush. Plans to depart around 7:30am to arrive at the office before 8:30am. Also considering buying a bike lock and installing a bike rack at the office. #commute #bicycle #work route
- [[podcasts]] : The user enjoys listening to podcasts during their commute, currently listening to "How I Built This" and feeling inspired by entrepreneurial stories, plans to continue listening while cycling to work #podcasts #entrepreneurship #commute entertainment
- [[indie rock]] : User has been listening to a lot of indie rock lately and is a fan of The Killers. Also familiar with The Strokes and Arctic Monkeys. Discovered The 1975 when they opened for The Killers at Red Rocks Amphitheater and really enjoyed their performance. #music #indie rock #The Killers
- [[film soundtracks]] : Recently listens to film soundtracks frequently, especially Hans Zimmer scores. Has a dedicated film soundtrack playlist on Spotify, adds new tracks weekly. Listening to film soundtracks helps maintain focus during work projects. #music #soundtracks #productivity
- [[anxiety and mental health management]] : User experiences work-triggered anxiety and panic attacks, and is exploring self-care strategies like mindful breathing and meditation while also struggling with self-consciousness and a desire to build confidence #anxiety #mental health #self-care
\end{lstlisting}

\medskip
\textbf{Step 3: Read Pages} $\to$ \texttt{bicycle commute.md}, \texttt{podcasts.md}, \texttt{indie rock.md}, \texttt{film soundtracks.md}, \texttt{anxiety and mental health management.md}\\
The agent reads all five pages to synthesize commute-appropriate activity suggestions:
\begin{itemize}[nosep, leftmargin=1.5em]
    \item \texttt{bicycle commute}: $\sim$35--45 min hybrid bike ride; audio-only activities suitable.
    \item \texttt{podcasts}: Currently listens to ``How I Built This''; wants to explore history/science genres (Hardcore History, StarTalk Radio, Radiolab).
    \item \texttt{indie rock}: Fan of The Killers, The Strokes, Arctic Monkeys, The 1975.
    \item \texttt{film soundtracks}: Hans Zimmer playlist on Spotify; helps focus.
    \item \texttt{anxiety management}: Mindful breathing to manage work-triggered anxiety.
\end{itemize}

\begin{lstlisting}[style=grayoutput]
[user/assistant/topic/bicycle commute.md]
# bicycle commute
- The user plans to buy a hybrid bicycle for weekday rush-hour commuting to avoid crowded Monday buses. Starting from 123 Oak St to 456 Main St. Prefers bike lanes and quiet streets, avoiding busy roads and steep hills. Planned route is approximately 2.6 miles, estimated 35-45 minutes during morning rush. Plans to depart around 7:30am to arrive at the office before 8:30am. Also considering buying a bike lock and installing a bike rack at the office.
- aliases: []
- tags: [commute, bicycle, work route]

## Planning to buy a bicycle for commuting and route planning
- category: experience
- detail: The user plans to buy a hybrid bicycle for weekday rush-hour commuting to avoid crowded Monday buses. Starting from 123 Oak St to 456 Main St. Prefers bike lanes and quiet streets, avoiding busy roads and steep hills. Planned route is approximately 2.6 miles, estimated 35-45 minutes during morning rush. Plans to depart around 7:30am to arrive at the office before 8:30am. Also considering buying a bike lock and installing a bike rack at the office.
- Page:
  - [[user/assistant/topic/daily routine and time management.md]]
  - [[user/assistant/topic/podcasts.md]]
  - [[user/assistant/place/123 oak st.md]]
  - [[user/assistant/place/456 main st.md]]

[user/assistant/topic/podcasts.md]
# podcasts
- The user enjoys listening to podcasts during their commute, currently listening to "How I Built This" and feeling inspired by entrepreneurial stories, plans to continue listening while cycling to work
- aliases: []
- tags: [podcasts, entrepreneurship, commute entertainment]

## Podcast listening preferences during commute
- category: preference
- detail: The user enjoys listening to podcasts during their commute, currently listening to "How I Built This" and feeling inspired by entrepreneurial stories. Plans to continue listening to podcasts while cycling to work.
- Page:
  - [[user/assistant/topic/bicycle commute.md]]
  - [[user/assistant/topic/daily routine and time management.md]]
  - [[user/assistant/place/123 oak st.md]]
  - [[user/assistant/place/456 main st.md]]

## Podcast listening preferences and commute habits
- category: preference
- detail: As of 2023-05-26, user listens to podcasts during their commute, which is about 40 minutes each way. They currently listen to true crime and self-improvement genres but want to branch out into other genres. They are particularly interested in history and science podcasts. Recommended history podcasts include Hardcore History, Lore, and The Dollop. Recommended science podcasts include StarTalk Radio, Radiolab, and Stuff You Should Know.
- Page:
  - [[user/assistant/topic/19th-century Korean history.md]]

[user/assistant/topic/indie rock.md]
# indie rock
- User has been listening to a lot of indie rock lately and is a fan of The Killers. Also familiar with The Strokes and Arctic Monkeys. Discovered The 1975 when they opened for The Killers at Red Rocks Amphitheater and really enjoyed their performance.
- aliases: []
- tags: [music, indie rock, The Killers, The 1975]

## Interest in indie rock music
- category: preference
- detail: User has been listening to a lot of indie rock lately and is a fan of The Killers. Also familiar with The Strokes and Arctic Monkeys. Discovered The 1975 when they opened for The Killers at Red Rocks Amphitheater and really enjoyed their performance.
- Page:
  - [[user/assistant/topic/live music and festivals.md]]
  - [[user/assistant/figure/Brandon Flowers.md]]
  - [[user/assistant/place/Denver.md]]
  - [[user/assistant/place/Nashville.md]]
  - [[user/assistant/place/Chicago.md]]
  - [[user/assistant/place/Red Rocks Amphitheater.md]]

[user/assistant/topic/film soundtracks.md]
# film soundtracks
- Recently listens to film soundtracks frequently, especially Hans Zimmer scores. Has a dedicated film soundtrack playlist on Spotify, adds new tracks weekly. Listening to film soundtracks helps maintain focus during work projects.
- aliases: []
- tags: [music, soundtracks, productivity]

## Film soundtrack preferences and listening habits
- category: preference
- detail: Recently listens to film soundtracks frequently, especially Hans Zimmer scores. Has a dedicated film soundtrack playlist on Spotify, adds new tracks weekly. Listening to film soundtracks helps maintain focus during work projects.
- Page:
  - [[user/assistant/topic/films.md]]
  - [[user/assistant/figure/family.md]]

[user/assistant/topic/anxiety and mental health management.md]
# anxiety and mental health management
- User experiences work-triggered anxiety and panic attacks, and is exploring self-care strategies like mindful breathing and meditation while also struggling with self-consciousness and a desire to build confidence
- aliases: []
- tags: [anxiety, mental health, self-care, panic attack, work stress]

## User's experience with anxiety and self-care strategies
- category: experience
- detail: User experiences anxiety and had a major panic attack at work around April 11, 2023 (about 6 weeks prior to May 23, 2023), triggered by work deadlines. To manage anxiety and prevent future attacks, user is exploring self-care activities. They are interested in mindful breathing, short meditation sessions, and progressive muscle relaxation. User plans to start using the Headspace app for guided recordings during lunch breaks. They are actively trying to prioritize well-being and make self-care a habit, especially when feeling overwhelmed with work.
- Page:
  - [[user/assistant/topic/work and project management.md]]

## Struggles with self-consciousness and desire to build confidence
- category: experience
- detail: As of 2023-05-27, the user struggles with being too self-conscious around new people and wants to be more comfortable in their own skin. They are actively working on building confidence and meaningful relationships. To step out of their comfort zone, they plan to attend social events and join a public speaking group.
\end{lstlisting}

\medskip
\textbf{Prediction:}\\
Listen to indie rock (The Killers, The Strokes, Arctic Monkeys, The 1975), film soundtracks (Hans Zimmer), or podcasts like ``How I Built This'' --- and try history/science podcasts (Hardcore History, StarTalk Radio, Radiolab). You could also practice mindful breathing during your $\sim$40-min bike commute to manage work anxiety. \checkmark
\end{tcolorbox}

\paragraph{Agent Autonomy.}
Crucially, the agent exercises full autonomy over its navigation decisions. At each step, the agent determines which dimension to enter, whether to drill deeper into a specific page, and whether to follow cross-dimensional associative links to related content. Each Think step constitutes genuine reasoning: the agent evaluates whether accumulated information is sufficient, whether additional context would improve its response, and whether related entries visible in the current page's link graph might be relevant. This autonomy means that \textbf{retrieval and reasoning are no longer separate stages but rather two facets of a single, unified cognitive process}. The agent does not first retrieve and then reason; it reasons \emph{through} the act of navigation itself, with each observation shaping subsequent exploration decisions. The ability to self-select the appropriate strategy---from single-step lookup to multi-step traversal---represents a form of metacognitive reasoning: before initiating navigation, the agent assesses the nature of the query and plans an efficient exploration path. This prevents both the under-exploration that plagues single-shot systems (missing relevant context) and the over-exploration that would result from always performing exhaustive multi-step traversals.

\end{document}